%% file: main.tex
\documentclass{article}

\usepackage[preprint]{corl_2026} % Uncomment for pre-prints (e.g., arxiv); This is like ``final'', but will remove the CORL footnote.

\usepackage{enumitem}
\usepackage{amsmath}
\usepackage{graphicx}
\usepackage{booktabs}
\usepackage{multirow}
\usepackage{wrapfig}
\usepackage{amssymb}
\usepackage[dvipsnames]{xcolor}
\usepackage{tabularx}
\usepackage{array}
\usepackage{enumitem}
\usepackage{algorithm}
\usepackage{algpseudocode}
\usepackage{multirow}

\title{Breaking Lock-In: Preserving Steerability under Low-Data VLA Post-Training}

\author{
  Suning Huang \hspace{0.12em}
  Jiaqi Shao \hspace{0.12em}
  Ke Wang \hspace{0.12em}
  Qianzhong Chen \hspace{0.12em}
  Jiankai Sun \hspace{0.12em}
  Yanjiang Guo \hspace{0.12em}\\[1ex]
  \textbf{Mac Schwager\textsuperscript{$\dagger$}} \hspace{0.12em}
  \textbf{Jeannette Bohg\textsuperscript{$\dagger$}} \\[1ex]
  Stanford University
}

\begin{document}
\maketitle

\makeatletter
\renewcommand{\@makefntext}[1]{\noindent #1}
\makeatother
\footnotetext[0]{\textsuperscript{$\dagger$} Equal advising. For any questions, please contact: \texttt{suning@stanford.edu}}

\newcommand{\model}{\texttt{DeLock}}

%%%%%%%%%%%%%%%%%%%%%%%%%%%%%%%%%%%%%%%%%%%%%%%%%
\input{texts/0_abstract}
\input{texts/1_introduction}
\input{texts/2_related_work}
\input{texts/3_method}
\input{texts/4_experiments}
\input{texts/5_conclusion}

\clearpage

\bibliography{main}  % .bib

\clearpage
\appendix
\section*{Appendix}
\input{texts/appendix}

\end{document}

%% file: texts/0_abstract.tex
\begin{abstract}
Have you ever post-trained a generalist vision-language-action~(VLA) policy on a small demonstration dataset, only to find that it stops responding to new instructions and is limited to behaviors observed during post-training? We identify this phenomenon as \emph{lock-in}: after low-data, supervised fine-tuning~(SFT), the policy becomes overly specialized to the post-training data and fails to generalize to novel instructions, manifesting as concept lock-in~(fixation on training objects/attributes) and spatial lock-in~(fixation on training spatial targets). Many existing remedies introduce additional supervision signals, such as those derived from foundation models or auxiliary objectives, or rely on augmented datasets to recover generalization. In this paper, we show that the policy's internal pre-trained knowledge is sufficient: {\model} mitigates lock-in by preserving visual grounding during post-training and applying test-time contrastive prompt guidance to steer the policy's denoising dynamics according to novel instructions. Across eight simulation and real-world evaluations, {\model} consistently outperforms strong baselines and matches or exceeds the performance of a state-of-the-art generalist policy post-trained with substantially more curated demonstrations. Experimental videos are available at \href{https://suninghuang19.github.io/delock\_page/}{DeLock}.

\end{abstract}

\keywords{Robot Foundation Models, Low-Data Post-Training, Contrastive Prompt Guidance} 

%% file: texts/1_introduction.tex
\section{Introduction}
\label{intro}

Generalist robot policies~(e.g., vision-language-action models, VLAs) have recently shown impressive generalization capabilities: they can perform a range of tasks in unseen environments and generalize across scene configurations, objects, and open-vocabulary language instructions~\citep{intelligence2025vision,khazatsky2024droid,o2024open,kim2024openvla,bjorck2025gr00t,pertsch2025fast}. Despite this breadth, generalist policies often must be adapted to perform effectively on downstream tasks or a specific robot embodiment, most commonly by post-training on tens to hundreds of hours of curated demonstrations for the target domain~\citep{team2025gemini,bousmalis2023robocat,black2024pi_0,guo2025ctrl,guo2026vlaw}. While prior work has shown that such post-training can yield robust policies, collecting large-scale demonstration corpora is expensive and often impractical. Consequently, in real deployments, these generalist policies are frequently post-trained in a low-data regime~(i.e., with limited demonstrations and narrow instruction coverage)~\citep{zang2025rlinf,li2025controlvla,cheng2025moe,guo2025improving,li2024evaluating}. This setting exposes a \emph{performance-generality dilemma} for standard supervised fine-tuning~(SFT): it can learn the target skill from limited demonstrations, yet often over-specializes to the post-training data distribution, making the learned skill hard to steer under novel instructions beyond the post-training conditions~\citep{shenfeld2025rl,fei2025libero,zhou2025libero,wortsman2022robust,jin2022dataless,yadav2025robust}.

\begin{wrapfigure}{r}{0.55\linewidth}
\centering
\includegraphics[width=0.95\linewidth]{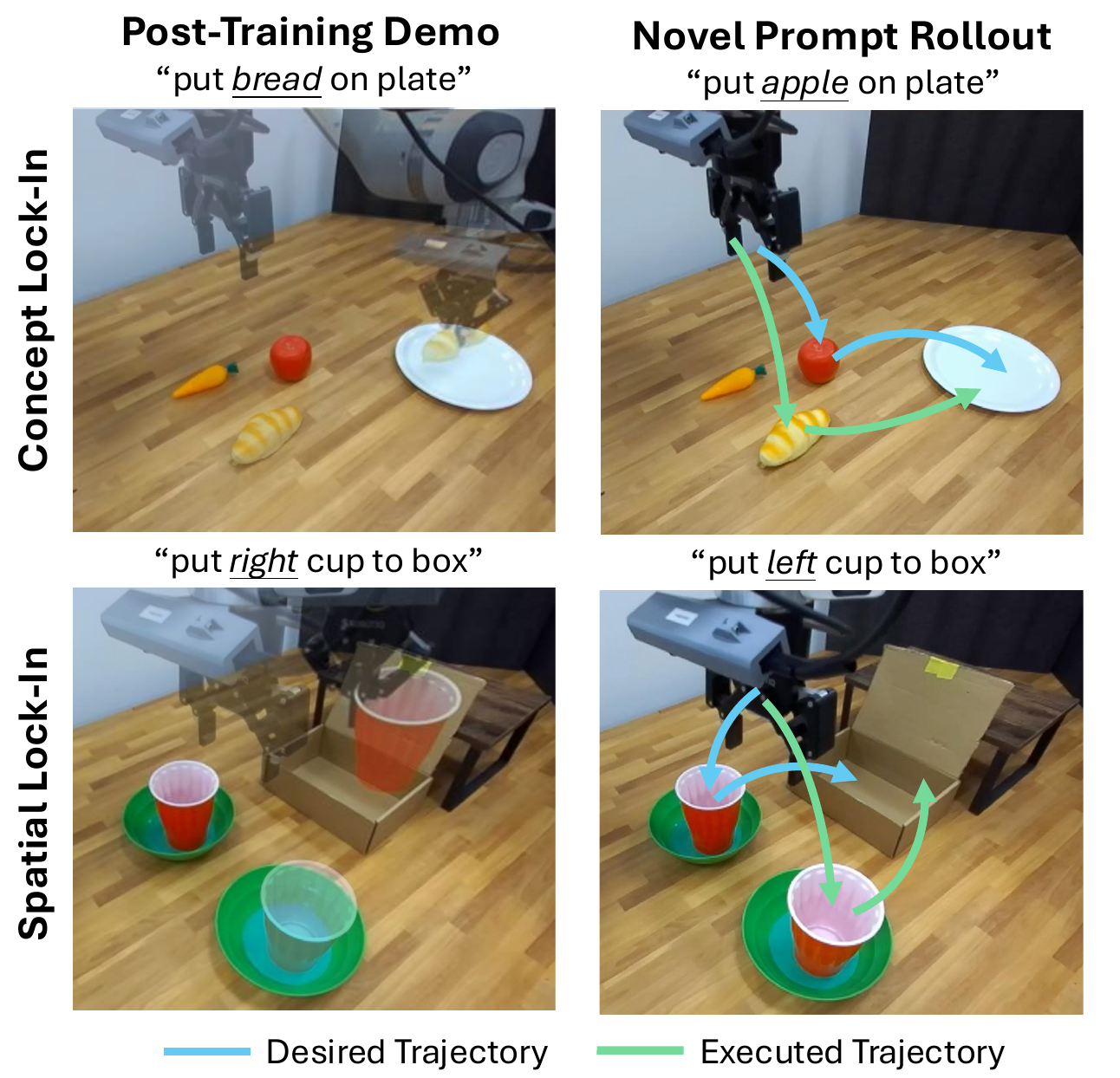}
\vspace{-0.8\baselineskip}
\caption{\textbf{Lock-In Failure Mode.} In low-data post-training, VLA policies can over-specialize into the training-demo distribution, becoming difficult to steer under novel prompts. We highlight concept lock-in under novel object concepts and spatial lock-in under novel spatial relations. Blue and green arrows denote desired and executed trajectories, respectively.}
\label{fig:teaser}
\vspace{-1.2\baselineskip}
\end{wrapfigure}

To mitigate this failure, existing literature often introduces additional supervision signals, such as those derived from foundation models~\citep{li2025controlvla,liu2026vls,chen2026steerable,li2025spatial,kachaev2510don} or auxiliary objectives~(e.g., VQA-style losses, dynamics prediction losses), or relies on augmented datasets~(e.g., human-robot co-training) to improve generalization~\citep{grover2025enhancing,lepert2025masquerade,punamiya2025egobridge}. However, these dependencies increase system complexity as well as training and deployment cost~\citep{bommasani2021opportunities,ma2024survey}, and they sidestep a key question: whether a VLA's internal pre-trained priors can be preserved and effectively leveraged to enable post-training generalization under realistic low-data settings, where exhaustive data coverage over concepts and spatial variants is unavailable. In this work, we formalize this common post-training failure mode in VLAs under the unified concept of \emph{lock-in}, as shown in Figure~\ref{fig:teaser}. Concretely, it manifests in two forms: \textit{concept lock-in}, where the policy collapses its language grounding and fixates on the training concept(s)~(e.g., specific object identities or attributes), executing the learned skill on them regardless of which concept the prompt specifies, and \textit{spatial lock-in}, where the policy remains anchored to the training spatial target(s)~(e.g., left vs.\ right, upper vs.\ lower) and continues acting toward those locations regardless of the spatial instruction in the prompt.

In this paper, we introduce {\model}, a simple yet effective framework for low-data post-training that preserves a generalist policy's concept and spatial generalization under narrow demonstrations and instruction coverage. {\model} combines two tightly coupled ingredients: (i) visual encoder weight-drift regularization during post-training to preserve pre-trained visual grounding and prevent representation collapse toward a narrow post-training distribution, and (ii) contrastive prompt guidance~(CPG), a classifier-free prompt-contrast rule~\citep{nguyen2024language,ban2024understanding,jang2023can,wan2024contrastive,jeong2025stylekeeper} applied at test time to steer action generation using the model's own conditional denoising dynamics. Specifically, the regularized model effectively improves steerability to novel concepts~(e.g., unseen object identities or attributes), while retaining the grounding signal needed for contrastive prompt guidance. In CPG, the trained prompt defines the negative condition, reflecting post-training bias toward training targets, while the novel prompt defines the positive condition. The difference between their denoising flows is then used as a test-time guidance signal to steer execution toward the novel instruction. To systematically evaluate concept- and spatial-reasoning generalization under low-data post-training, we further develop a dedicated benchmark spanning both simulation and real-world settings.

In summary, our contributions are threefold: 
(1) We formalize \emph{lock-in} as a common low-data post-training failure mode in VLAs, and distinguish two forms: concept lock-in and spatial lock-in. These failure modes manifest as broken language grounding and degraded instruction-following generalization beyond the post-training distribution. 
(2) We propose {\model}, a simple yet effective framework for low-data post-training that jointly preserves and exploits a VLA's internal pre-trained priors: visual encoder weight-drift regularization preserves pre-trained grounding during post-training, while contrastive prompt guidance~(CPG) leverages the preserved grounding to steer the policy at test time.
(3) Since no existing benchmark directly probes these lock-in failures, we build a new evaluation suite with four LIBERO-based~\citep{liu2023libero} simulation tasks and four real-world tasks on the DROID setup~\citep{khazatsky2024droid}, and show that {\model} consistently outperforms strong baselines, matching or exceeding the performance of a state-of-the-art generalist policy~(e.g., $\pi_{0.5}\text{-DROID}$~\citep{intelligence2025vision}) post-trained with substantially more curated demonstrations.

%% file: texts/2_related_work.tex
\section{Related Work}
\label{related_work}

\paragraph{Post-Training Generalist Policies in Low-Data Regime.}
Generalist policies can solve diverse tasks, but achieving strong performance on a target skill typically requires post-training on task demonstrations~\citep{zitkovich2023rt,anil2023palm,chen2025sarm}. Prior adaptation paradigms include supervised fine-tuning~\citep{black2023zero,team2024octo}, in-context test-time improvement~\citep{fu2024context,sharma2023lossless}, and RL-based adaptation~\citep{pan2026sop,xiang2025parallels,huang2024mentor,hu2022lora,li2025simplevla}; however, in the common low-data setting, post-training often causes the policy to over-specialize to dataset-specific biases, making the learned skill difficult to steer under novel instructions~\citep{xing2025shortcut,fei2025libero}. In contrast, {\model} preserves pre-trained visual grounding during low-data post-training and enables contrast prompt guidance for generalization beyond the post-training instructions.

\paragraph{Inference-Time Guidance for Generative Policies.}
After post-training, policy behavior can still be shaped at rollout time without modifying weights~\citep{wang2024poco,cao2025compose}. Prior work achieves such steering by introducing external guidance sources, such as learned dynamics models that impose predictive constraints~\citep{du2025dynaguide,sun2025latent,huang2025particleformer}, value signals that bias sampling~\citep{koulischer2024dynamic,nakamoto2024steering,wagenmaker2025steering}, or human/VLM feedback that selects or filters candidate behaviors~\citep{xu2023xskill,liu2026vls,li2025towards}. They introduce additional training/deployment dependencies and can inherit biases from auxiliary objectives. {\model} instead steers via the policy's own pre-trained grounding, using prompt contrast as an internal guidance signal without auxiliary models or human/VLM intervention.

%% file: texts/3_method.tex
\section{Method}
\label{method}

\begin{figure}[ht]
\centering
\includegraphics[width=1\linewidth, trim= 0cm 0cm 0cm 0cm, clip]{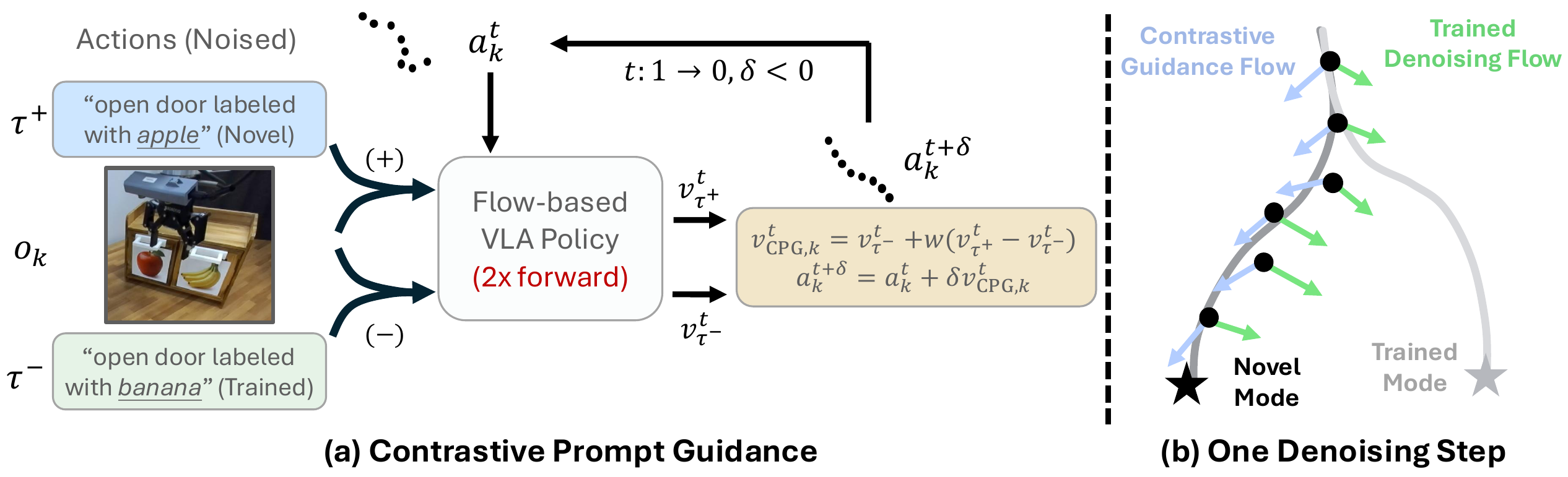}
\vspace{-1.5em}
\caption{\textbf{Test-Time Contrastive Prompt Guidance.} {\model} uses contrastive prompt guidance at inference time to steer a post-trained flow-based VLA toward novel instructions. 
\textbf{(a)} At rollout step $k$, the policy conditions on observation $o_k$ and iteratively denoises a noised action chunk $a_k^{t}$ from $t{=}1$ to $0$~(with $\delta<0$). At each denoising step, the same policy is \textbf{forwarded twice} with a \textit{positive}~(novel) prompt $\tau^{+}$ and a \textit{negative}~(trained) prompt $\tau^{-}$ to obtain $v_{\tau^{+}}^{t}$ and $v_{\tau^{-}}^{t}$, which are combined as $v_{\mathrm{CPG}}^{t}=v_{\tau^{-}}^{t}+w\!\left(v_{\tau^{+}}^{t}-v_{\tau^{-}}^{t}\right)$ to update $a_k^{t+\delta}=a_k^{t}+\delta\, v_{\mathrm{CPG}}^{t}$. 
\textbf{(b)} An example of one denoising step: the prompt contrast steers the denoising trajectory away from the trained mode toward a mode aligned with the novel instruction. Detailed pseudocode is provided in Appendix~\ref{app:pseudo_code}.}
\label{fig:method}
\end{figure}

\subsection{Problem Formulation}
\label{method:formulation}

Given a pre-trained generalist policy, our goal is to post-train it on a small, narrowly covered demonstration set for a target skill while avoiding lock-in.

% : after standard supervised fine-tuning~(SFT), the policy becomes proficient at the demonstrated interaction pattern but fails to follow \emph{novel} instructions that change the intended concept~(object identity/attribute) or spatial target within the same scene.

\paragraph{Standard SFT.}
A VLA policy $\pi_\theta(a\mid o,\tau)$ maps an observation $o\in\mathcal{O}$ and instruction $\tau\in\mathcal{T}$ to an action distribution over $a\in\mathcal{A}$. Pre-training yields $\pi_{\theta_{\mathrm{pre}}}$ by minimizing $\mathcal{L}_{\mathrm{pre}}(\theta;\mathcal{D}_{\mathrm{pre}})$ on a large-scale dataset $\mathcal{D}_{\mathrm{pre}}=\{(o_i,a_i,\tau_i)\}_{i=1}^{N_{\mathrm{pre}}}$, giving $\theta_{\mathrm{pre}}$. In low-data post-training, we are given $\mathcal{D}_\star=\{(o_j,a_j,\tau_j)\}_{j=1}^{N_\star}$ with $N_\star\ll N_{\mathrm{pre}}$ and narrow instruction coverage $\tau_j\in\mathcal{T}_\star\subset\mathcal{T}$ (often from a fixed scene with restricted concept/spatial variants). Standard SFT initializes from $\theta_{\mathrm{pre}}$ and minimizes the behavioral-cloning loss $\mathcal{L}_{\mathrm{BC}}(\theta;\mathcal{D}_\star)=-\mathbb{E}_{(o,a,\tau)\sim\mathcal{D}_\star}\!\left[\log \pi_\theta(a\mid o,\tau)\right]$, producing $\pi_{\theta_{\mathrm{ft}}}$.

\paragraph{Concept Lock-In vs. Spatial Lock-In.}
Lock-in manifests when $\pi_{\theta_{\mathrm{ft}}}$ is insensitive to instructions that were not covered in $\mathcal{D}_\star$, despite having learned the underlying skill and having seen the underlying concepts in the pretraining data. As illustrated in Figure~\ref{fig:teaser}, we categorize this into two types. Concept lock-in occurs when post-training demonstrations involve only a subset of concepts~(e.g., only ``pick bread'' among multiple randomly placed objects); at test time, the policy remains fixated on the training concept even when the prompt specifies a different one~(e.g., ``pick apple''). Spatial lock-in occurs when demonstrations only cover a specific spatial target~(e.g., always ``pick right cup'' when two cups are present), leading the policy to ignore alternative spatial cues in the prompt~(e.g., ``pick left cup''). Avoiding such lock-in \textit{without} exhaustively collecting demonstrations for every concept and spatial variant is the central challenge we address.

\subsection{Visual Encoder Weight-Drift Regularization for Knowledge Preservation}
\label{method:vit_reg}

\paragraph{Visual Grounding as a Bottleneck for Avoiding Lock-In.}
Most modern generalist VLA policies factor into three components: a visual encoder $v_{\theta_v}$ that extracts visual features from the observation $o$, a language backbone $l_{\theta_\ell}$ that conditions these features on the instruction $\tau$, and an action expert $a_{\theta_a}$ that decodes robot actions. We parameterize the policy as $a \sim \pi_\theta(\cdot\mid o,\tau)$, where the policy is instantiated as $a_{\theta_a}\!\big(l_{\theta_\ell}(\tau,\; v_{\theta_v}(o))\big)$. In standard low-data SFT, practitioners often use LoRA to prevent over-updating the language backbone and action expert, but still allow the visual encoder to update freely. Since instruction grounding to scene concepts and spatial relations is mediated by the visual features $v_{\theta_v}(o)$, drift in the visual encoder can directly impair this grounding, and downstream~(even general) language/action modules have limited ability to recover it. We therefore seek to preserve the pre-trained visual priors while still allowing task-specific adaptation in the downstream modules.

\paragraph{L2 Regularization on Visual Encoder Drift.}
Let $\theta_v^{\mathrm{pre}}$ denote the visual encoder parameters of the pre-trained model and $\theta_v$ the visual encoder parameters during post-training. {\model} augments the SFT objective with an $L_2$ penalty that discourages $\theta_v$ from drifting far from $\theta_v^{\mathrm{pre}}$:
\begin{equation}
\mathcal{L}_{\model}(\theta;\mathcal{D}_\star)
=
\mathcal{L}_{\mathrm{BC}}(\theta;\mathcal{D}_\star)
+\lambda \big\lVert \theta_v-\theta_{v}^{\mathrm{pre}} \big\rVert_2^2,
\label{eq:vreg}
\end{equation}
\noindent where $\lambda$ controls the regularization strength. We regularize the visual encoder parameters $\theta_v$ during post-training and adapt $l_{\theta_\ell}$ and $a_{\theta_a}$ as in standard low-data SFT~(e.g., via LoRA), preserving pre-trained visual grounding while acquiring the target skill.

\subsection{Contrastive Prompt Guidance at Test Time}
\label{method:cpg}

Building on the visually preserved model, we introduce contrastive prompt guidance~(CPG), a test-time prompt-contrast rule that leverages the policy's retained grounding knowledge to steer its conditional denoising dynamics and better align execution with novel instructions, as shown in Figure~\ref{fig:method}.

\paragraph{Flow-Matching Action Generation.}
Current state-of-the-art VLA policies commonly parameterize action generation as a conditional denoising process~\citep{intelligence2025vision,bjorck2025gr00t,black2024pi_0}. We define $t\in[0,1]$ with $t=1$ corresponding to pure noise and $t=0$ to the target action distribution. Let $a_{k}^{t}$ denote the action chunk at rollout step $k$, flow time $t$, initialized as $a_{k}^{1}\sim\mathcal{N}(0,I)$. Conditioned on observation $o_k$ and instruction $\tau$, the policy predicts a denoising vector field $v_\theta(o_k,\tau,t)$, and generates actions by integrating this field via an Euler update rule:
$
a_{k}^{t+\delta}=a_{k}^{t}+\delta\, v_\theta(o_k,\tau, t),
$
where $\delta<0$ and we iterate from $t=1$ to $t=0$ to obtain the final action $a_{k}=a_{k}^{0}$.

\paragraph{Prompt-Contrast Steering Rule.}
{\model} steers the denoising dynamics by contrasting a \emph{positive} prompt $\tau^{+}$, corresponding to the novel instruction, with a \emph{negative} prompt $\tau^{-}$. In this work, we use the trained instruction as the negative prompt, as it captures the post-training bias toward the training targets. Leveraging the retained grounding from Section~\ref{method:vit_reg}, we use the prompt-induced contrast between the two conditional vector fields as an inference-time guidance signal. Concretely, we define the guided field as
\begin{equation}
v_{\mathrm{CPG},k}^{t}
=
v_\theta(o_{k},\tau^{-},t)
+
w\Big(v_\theta(o_{k},\tau^{+},t)-v_\theta(o_{k},\tau^{-},t)\Big),
\label{eq:flow_cpg}
\end{equation}
\noindent where $w\ge 0$ is a tunable guidance scale, and apply it by replacing the Euler update with
$
a_{k}^{t+\delta}=a_{k}^{t}+\delta\, v_{\mathrm{CPG},k}^{t}.
$
This amplifies the contrastive signal $v_{\tau^{+}} - v_{\tau^{-}}$, strengthening instruction-relevant directions while suppressing shared bias, so the guidance remains effective even when $v_{\tau^{+}}$ itself is already overfitting to training targets after low-data post-training.

%% file: texts/4_experiments.tex
\section{Experiments}
\label{exp}

In this section, we empirically evaluate {\model} and answer three questions: (1) How can we design evaluation benchmarks that systematically probe the lock-in failure introduced above? (2) What are the representation- and denoising-dynamics signatures of lock-in, and how does {\model} mitigate them? (3) How does {\model} compare with strong low-data post-training baselines and with generalist policies post-trained on substantially larger curated demonstration sets? Unless otherwise stated, all experiments start from a pre-trained flow-based VLA model $\pi_{0.5}\text{-BASE}$~\citep{intelligence2025vision} and apply low-data post-training under the protocols described below. Training details are provided in Appendix~\ref{app:post_train}.

\subsection{Benchmarking Lock-In under Low-Data Post-Training}
\label{benchmark}

\begin{figure}[ht]
\centering
\includegraphics[width=1\linewidth, trim= 0cm 0cm 0cm 0cm, clip]{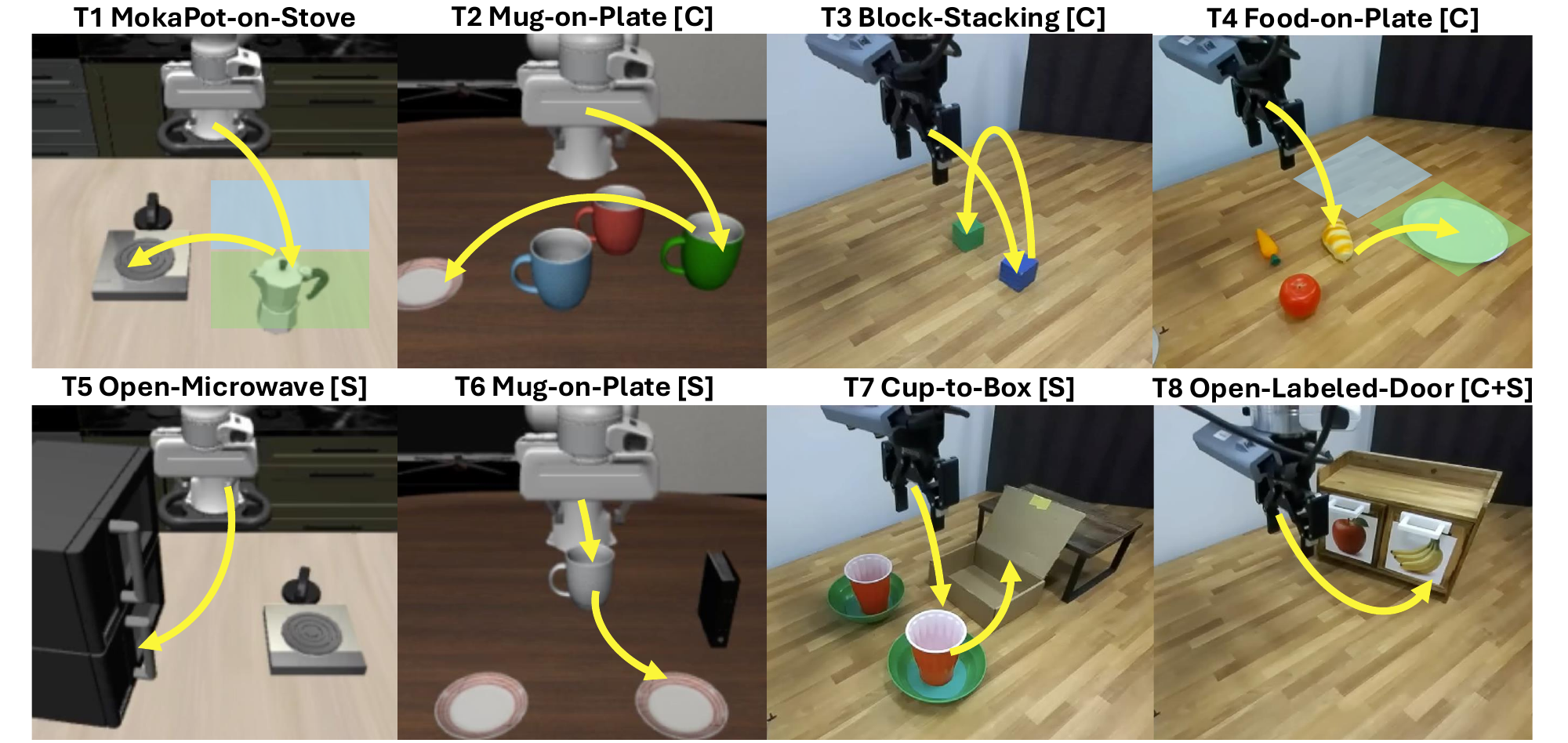}
\vspace{-1.5em}
\caption{\textbf{Lock-In Failure Evaluation Benchmark.} Our 8-task suite spans four LIBERO simulation tasks and four real-world DROID tasks. Labels \textbf{[C]} and \textbf{[S]} denote concept- and spatial-lock-in probes, respectively. Yellow arrows illustrate the manipulation pattern demonstrated during post-training. In tasks with shaded regions, we additionally evaluate OOD location shift: the green-shaded region indicates the object placement distribution in post-training demonstrations, and the blue-shaded region indicates the shifted placement distribution used at evaluation.}
\label{fig:benchmark_suite}
\end{figure}

A key challenge in studying lock-in is that many existing benchmarks primarily test how well a post-trained policy performs the trained task, often using the same instructions as in post-training~\citep{liu2023libero,li2025controlvla,chen2025sarm}. While these benchmarks do evaluate generalization, their scope is typically limited to visual or spatial distribution shifts---such as out-of-distribution~(OOD) object locations, background variations, or the presence of irrelevant distractors---rather than whether the policy remains responsive to changes in the instruction itself. As a result, strong performance under visual diversity can mask a policy's inability to \emph{re-steer} a learned skill when only the instruction changes.

To address this gap, we introduce an 8-task evaluation suite spanning both simulation and the real world, as shown in Figure~\ref{fig:benchmark_suite}. With one exception that focuses purely on a standard OOD location shift~(\textsc{MokaPot-on-Stove}), all other tasks are designed as paired lock-in probes: post-training demonstrations cover a restricted set of concept and/or spatial variants, while evaluation keeps the scene fixed and changes only the concept token or the spatial token in the instruction. The suite includes four LIBERO-based simulation tasks~(100 demonstrations per task) and four real-world tasks on the DROID setup~(80 demonstrations per task). Table~\ref{tab:benchmark_tasks} details the specific contrast between post-training and novel evaluation prompts.

\begin{table}[t]
\caption{\textbf{Instruction-level Shifts for Probing Lock-In Failure.} We show the post-training prompts and the novel evaluation prompts for each task. \emph{Italicized} tokens mark the instruction components varied between post-training and evaluation. For prompts with brackets, one listed variant is sampled per trial. Detailed task designs are provided in Appendix~\ref{app:exp_design}.}
\label{tab:benchmark_tasks}
\centering
\footnotesize
\setlength{\tabcolsep}{3.2pt}
\renewcommand{\arraystretch}{1.0}
\resizebox{0.97\linewidth}{!}{%
\begin{tabular}{@{}llll@{}}
\toprule
Task ID & Task Name & Post-Training Prompt & Novel Prompt for Evaluation \\
\midrule
T1 & MokaPot-on-Stove & \texttt{put moka pot on stove} & -- \\
T2 & Mug-on-Plate [C] & \texttt{put \textit{\underline{green}} mug on plate} & \texttt{put \{\textit{\underline{red}}, \textit{\underline{blue}}\} mug on plate} \\
T3 & Block-Stacking [C] & \texttt{stack \textit{\underline{blue}} block on \textit{\underline{green}} block} & \texttt{stack \textit{\underline{green}} block on \textit{\underline{blue}} block} \\
T4 & Food-on-Plate [C] & \texttt{put \textit{\underline{bread}} on plate} & \texttt{put \{\textit{\underline{apple}}, \textit{\underline{carrot}}\} on plate} \\
T5 & Open-Microwave [S] & \texttt{open \textit{\underline{lower}} microwave} & \texttt{open \textit{\underline{upper}} microwave} \\
T6 & Mug-on-Plate [S] & \texttt{put mug on \textit{\underline{left}} plate} & \texttt{put mug on \textit{\underline{right}} plate} \\
T7 & Cup-to-Box [S] & \texttt{put \textit{\underline{right}} cup to box} & \texttt{put \textit{\underline{left}} cup to box} \\
T8 & Open-Labeled-Door [C+S] & \texttt{open door labeled with \textit{\underline{banana}}} & \texttt{open door labeled with \textit{\underline{apple}}} \\
\bottomrule
\end{tabular}%
}
\end{table}

\subsection{Mechanistic Analysis: Signatures of Lock-In}
\label{exp:mechanism}

\begin{figure}[ht]
\centering
\includegraphics[width=1\linewidth, trim= 0cm 0cm 0cm 0cm, clip]{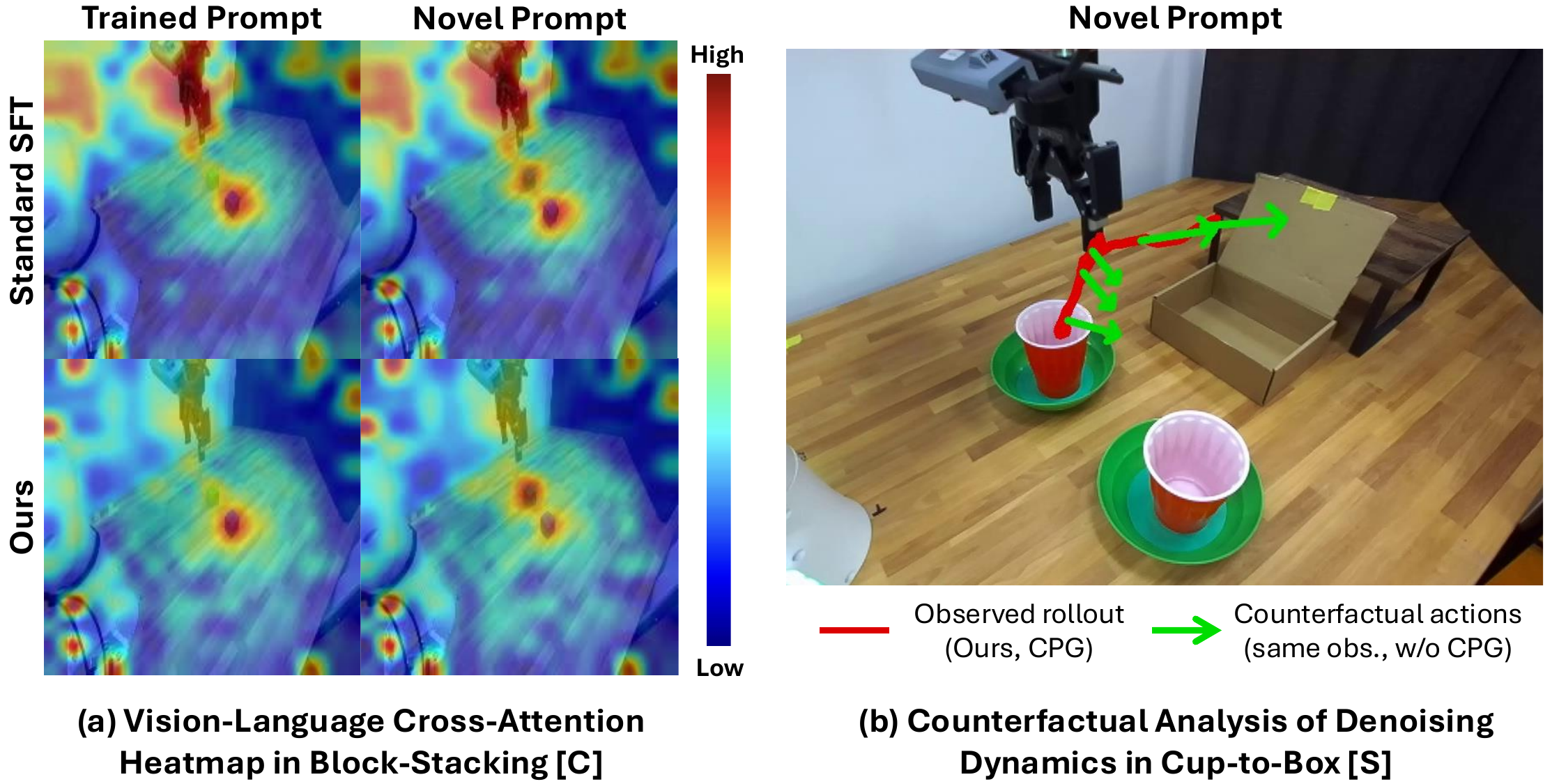}
\vspace{-1.5em}
\caption{\textbf{Qualitative Evaluation of Lock-In Failure.}
(a) \textsc{Block-Stacking}~[C], from ``stack blue block on green block'' to ``stack green block on blue block''. Standard SFT shows weak prompt-conditioned attention shift, while {\model} exhibits clearer instruction-aligned attention reallocation.
(b) \textsc{Cup-to-Box}~[S], evaluated on the novel prompt ``put left cup to box''. The red curve shows the observed rollout with {\model}~(CPG enabled), while the green arrows show a counterfactual replay on the same observations without CPG, which remains biased toward the trained target~(right cup).}
\label{fig:qualitative}
\vspace{-1em}
\end{figure}

We qualitatively analyze lock-in through two complementary lenses: (i) representation-level grounding, and (ii) generation-dynamics bias during action denoising, as illustrated in Figure~\ref{fig:qualitative}.

\paragraph{Collapsed Visual Grounding Induces Concept Lock-In.}
We use \textsc{Block-Stacking}~[C] as a controlled probe for concept lock-in: post-training demonstrations only contain ``stack blue block on green block'' while evaluation reverses the concept order to ``stack green block on blue block''. Although the required stacking skill is unchanged, standard SFT repeats the training behavior, whereas {\model} follows the new instruction and succeeds. To examine the representation-level cause, we visualize vision-language cross-attention in the PaliGemma~\citep{beyer2024paligemma} backbone using instruction tokens as queries and image patches as keys~(Figure~\ref{fig:qualitative}(a)). Standard SFT shows a collapsed attention pattern, continuing to focus on the blue block regardless of the prompt. In contrast, {\model} exhibits a clear prompt-conditioned shift in attention between the blue and green blocks as their instructed roles are swapped, consistent with preserved visual grounding under low-data post-training.

\paragraph{Spatial Lock-In as Biased Denoising Dynamics.}
We analyze spatial lock-in on \textsc{Cup-to-Box}~[S], where post-training demonstrations only cover picking the right cup and moving it to the box. At evaluation, the instruction is changed to ``put left cup to box''. Standard SFT remains biased toward the trained rightward behavior, whereas {\model} successfully reuses the learned pick-and-place skill and redirects it to the left cup. To understand the role of CPG, we perform a counterfactual rollout analysis~(Figure~\ref{fig:qualitative}(b)). Using the same observation sequence, removing CPG causes the policy to continue producing actions biased toward the trained target~(green arrows). In contrast, CPG adds a prompt-contrast term that strengthens the instruction-change direction and steers the denoising trajectory toward the new spatial target.

\subsection{Generalizing Beyond Post-Training: OOD Configurations and Novel Instructions}
\label{exp:quant}

We quantitatively evaluate \textbf{OOD performance}, comparing {\model} against (i) a strong low-data post-training baseline, (ii) a large-scale post-trained generalist reference, and (iii) targeted ablations to isolate the contributions of visual encoder regularization and contrastive prompt guidance. Please refer to Appendix~\ref{app:exp_result} for more experimental results.

\paragraph{Baselines.}
(1) RETAIN, a strong low-data post-training method that employs weight-space interpolation between the task-specific and pre-trained models to mitigate catastrophic forgetting; and (2) $\pi_{0.5}\text{-DROID}$, a state-of-the-art generalist VLA post-trained on the large-scale, curated DROID dataset, which serves as a high-resource reference rather than a low-data baseline, representing the upper bound achieved through massive data curation. 

\paragraph{Ablations.}
(1) {\model} w/o CPG, which removes the contrastive guidance at inference to assess the impact of raw policy grounding; (2) {\model} w/o Vis-Reg, which fine-tunes the visual encoder without regularization to evaluate the necessity of grounding preservation; and (3) {\model} w/ Frozen-Vis, which fixes the visual encoder during training as a rigid alternative to our regularized adaptation.

\begin{table}[t]
\caption{\textbf{Generalization under OOD Locations and Novel Instructions~(20 trials per task).} We report success counts for spatial generalization under OOD location shifts~(T1, T4) and for generalization to novel prompts across the seven lock-in probes. Task IDs follow Table~\ref{tab:benchmark_tasks}; $\pi_{0.5}\text{-DROID}$ is evaluated exclusively in real-world; full in-distribution results are provided in Appendix~\ref{app:exp_result}.}
\label{tab:quant_main}
\centering
\footnotesize
\setlength{\tabcolsep}{4pt}
\renewcommand{\arraystretch}{1.0}
\hspace*{-0.02\textwidth}%
\begin{minipage}{1.04\textwidth}
\centering
\makebox[\textwidth][c]{%
\begin{tabular}{@{}lcc|ccccccc@{}}
\toprule
\multirow{2}{*}[-0.5ex]{Method} & \multicolumn{2}{c|}{\textbf{OOD Locations}} & \multicolumn{7}{c}{\textbf{OOD Instructions~(Novel Prompts)}} \\
\cmidrule(r){2-3}\cmidrule(l){4-10}
& T1 & T4 [C] & T2 [C] & T3 [C] & T4 [C] & T5 [S] & T6 [S] & T7 [S] & T8 [C+S] \\
\midrule
RETAIN & 10/20 & 14/20 & 0/20 & 6/20 & 3/20 & 0/20 & 0/20 & 2/20 & 1/20 \\
$\pi_{0.5}\text{-DROID}$ & -- & \textbf{18}/20 & -- & 18/20 & \textbf{18}/20 & -- & -- & 11/20 & 0/20 \\
{\model} w/o CPG & \textbf{16}/20 & 16/20 & 17/20 & 18/20 & 15/20 & 0/20 & 0/20 & 0/20 & 0/20 \\
{\model} w/o Vis-Reg & 4/20 & 9/20 & 9/20 & 7/20 & 2/20 & 0/20 & 0/20 & 0/20 & 0/20 \\
{\model} w/ Frozen-Vis & 7/20 & 13/20 & 16/20 & 14/20 & 13/20 & 2/20 & 11/20 & 8/20 & 4/20 \\
{\model} & \textbf{16}/20 & 15/20 & \textbf{19}/20 & \textbf{19}/20 & 17/20 & \textbf{11}/20 & \textbf{13}/20 & \textbf{14}/20 & \textbf{13}/20 \\
\bottomrule
\end{tabular}%
}
\end{minipage}
\end{table}

\begin{figure}[ht]
\centering
\includegraphics[width=1\linewidth, trim= 0cm 0cm 0cm 0cm, clip]{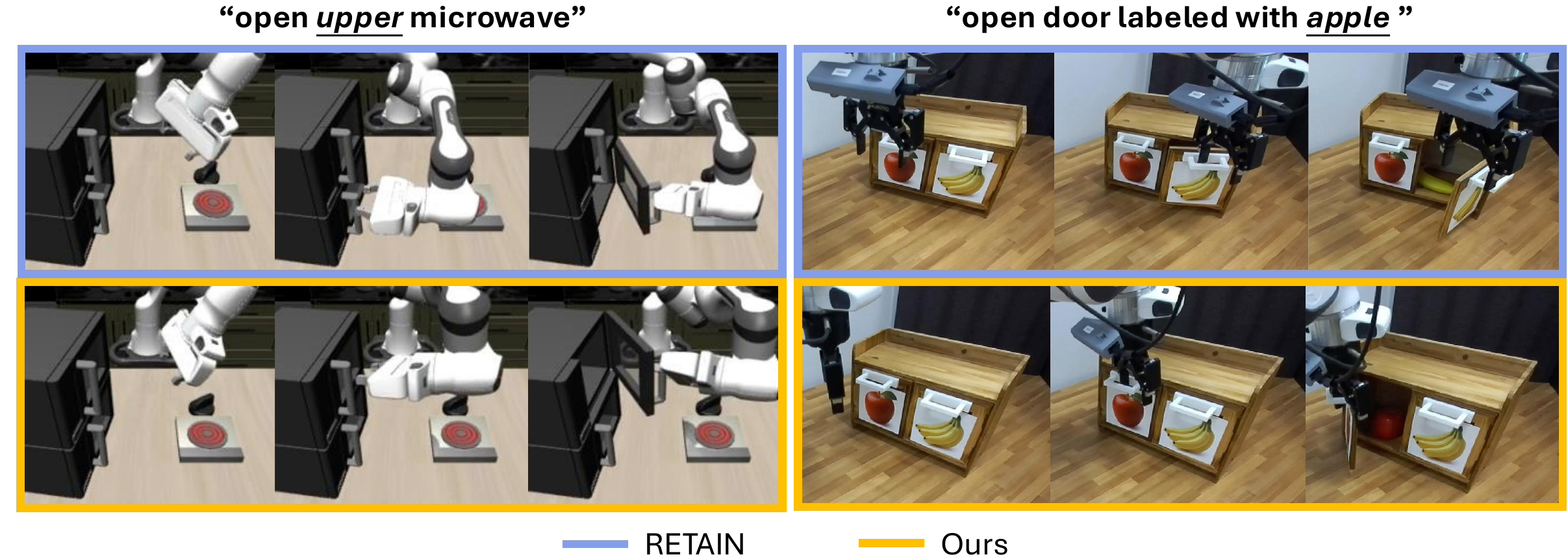}
\vspace{-1.5em}
\caption{\textbf{Novel-Prompt Rollouts on Articulated Tasks.} We compare {\model} and RETAIN on the two challenging tasks involving articulated objects: \textsc{Open-Microwave}~[S] and \textsc{Open-Labeled-Door}~[C+S]. Under novel prompts, RETAIN largely repeats the post-training trajectory and fails to follow the changed spatial/concept specification, whereas {\model} successfully re-steers the learned skill to follow the new instruction.}
\label{fig:exp_demo}
\vspace{-1em}
\end{figure}

Table~\ref{tab:quant_main} summarizes the success rates under both OOD spatial configurations and novel-instruction generalization. Overall, {\model} demonstrates superior across-the-board performance: it maintains high reliability under scene perturbations and significantly outperforms baselines on both concept- and spatial-lock-in probes. Figure~\ref{fig:exp_demo} qualitatively compares {\model} and RETAIN in their ability to follow novel prompts across two challenging articulated tasks.
% These results validate the synergy between preserving pre-trained visual grounding and employing test-time prompt-contrast steering.

Among the baselines, RETAIN exhibits moderate robustness to OOD location shifts, suggesting that parameter merging can partially mitigate sensitivity to visual-spatial distribution shifts. However, it fails to generalize across all novel-prompt tasks, indicating that simple weight-space interpolation is insufficient to overcome behavioral lock-in under instruction changes. In contrast, while $\pi_{0.5}\text{-DROID}$ performs well on OOD locations and some simpler concept shifts, its performance depends on an extensive post-training distribution that is often unavailable in specialized domains. Moreover, its performance drops substantially on spatial reasoning tasks such as \textsc{Cup-to-Box}~[S], where its behavior resembles stochastic switching between targets, and it fails entirely on the more fine-grained \textsc{Open-Labeled-Door}~[C+S] task. Together, these results suggest that even large-scale generalists struggle when adaptation requires fine-grained instruction following beyond dominant post-training patterns, highlighting the necessity for effective task-specific adaptation in the low-data regime.

Our ablation studies further disentangle the contributions of each component. Removing visual encoder regularization~(\model~w/o Vis-Reg) substantially degrades performance under both OOD location shifts and novel-prompt settings, confirming that preserving pre-trained grounding is critical for robust downstream generalization. However, preservation alone is not sufficient: removing CPG~(\model~w/o CPG) retains some success on concept probes but fails on all spatial-lock-in tasks, showing that prompt-contrast steering is essential for correcting training-induced spatial bias during rollout. Finally, fully freezing the visual encoder~(\model~w/ Frozen-Vis) yields non-trivial performance but consistently underperforms {\model}, underscoring the benefit of controlled, regularized adaptation over rigid parameter freezing. Together, these results show that {\model}'s effectiveness arises from the complementary roles of grounding preservation and test-time steering.

%% file: texts/5_conclusion.tex
\section{Conclusion}

In this paper, we formalize \emph{lock-in} as a common failure mode of low-data post-training in generalist VLA policies, and distinguish two forms: concept lock-in and spatial lock-in. We show that, under limited post-training data, policies can over-specialize to demonstration biases and lose the ability to re-steer learned skills under novel instructions. To address this, we introduce {\model}, which combines visual encoder weight-drift regularization to preserve pre-trained grounding with test-time contrastive prompt guidance to steer execution. Across both simulated and real-world tasks, our results show that this preserve-and-steer design enables effective instruction-conditioned OOD generalization without relying on large-scale post-training data or external sources of supervision.

\paragraph{Limitations.} Our study focuses on lock-in under relatively controlled low-data post-training settings, where the trained and novel prompts are specified a priori for contrastive guidance, and it remains to be seen how the same mechanisms scale to broader instruction distributions, longer-horizon tasks, and more open-ended real-world environments. In addition, our current visual encoder regularization and guidance scale design are relatively simple; more adaptive regularization schemes and context-dependent guidance strategies may further improve the trade-off between grounding preservation and task-specific adaptation.

%% file: texts/appendix.tex
\section{Pseudocode for Training and Inference}
\label{app:pseudo_code}

For completeness, we provide pseudocode for the two core components of {\model}. Algorithm~\ref{alg:vit_reg} describes the training-time procedure, where the visual encoder is regularized toward its pre-trained parameters while the language backbone and action expert are adapted via LoRA. Algorithm~\ref{alg:cpg} describes the test-time contrastive prompt guidance rule used for action denoising under novel instructions.

\begin{algorithm}[]
\caption{Low-Data Post-Training with Visual Encoder Weight-Drift Regularization}
\label{alg:vit_reg}
\begin{algorithmic}[1]
\Require pre-trained policy parameters $\theta_v^{\mathrm{pre}}, \theta_\ell^{\mathrm{pre}}, \theta_a^{\mathrm{pre}}$, low-data post-training set $\mathcal{D}_\star$, regularization weight $\lambda$
\Ensure post-trained parameters $\theta_v, \theta_\ell, \theta_a$

\State Initialize visual encoder parameters $\theta_v \gets \theta_v^{\mathrm{pre}}$
\State Initialize language backbone and action expert from pre-trained weights
\State Insert LoRA adapters into $l_{\theta_\ell}$ and $a_{\theta_a}$
\State Freeze base parameters of $l_{\theta_\ell}$ and $a_{\theta_a}$ except LoRA parameters
\State Keep a frozen copy of pre-trained visual parameters $\theta_v^{\mathrm{pre}}$ as reference

\While{not converged}
    \State Sample minibatch $(o,a,\tau) \sim \mathcal{D}_\star$
    \State Compute policy output:
    \State \hspace{1.5em} $\hat{a} \sim \pi_\theta(\cdot \mid o,\tau)
    = a_{\theta_a}\!\big(l_{\theta_\ell}(\tau, v_{\theta_v}(o))\big)$
    \State Compute behavioral cloning loss:
    \State \hspace{1.5em} $\mathcal{L}_{\mathrm{BC}} \gets -\mathbb{E}_{(o,a,\tau)\sim\mathcal{D}_\star}\!\left[\log \pi_\theta(a\mid o,\tau)\right]$
    \State Compute visual encoder drift penalty:
    \State \hspace{1.5em} $\mathcal{L}_{\mathrm{reg}} \gets \lambda \lVert \theta_v - \theta_v^{\mathrm{pre}} \rVert_2^2$
    \State Form total loss:
    \State \hspace{1.5em} $\mathcal{L}_{\model} \gets \mathcal{L}_{\mathrm{BC}} + \mathcal{L}_{\mathrm{reg}}$
    \State Update $\theta_v$ using $\nabla_{\theta_v}\mathcal{L}_{\model}$
    \State Update only the LoRA parameters in $l_{\theta_\ell}$ and $a_{\theta_a}$
\EndWhile

\State \Return $\theta_v, \theta_\ell, \theta_a$
\end{algorithmic}
\end{algorithm}

We next describe the inference-time steering rule applied to the post-trained policy.

\begin{algorithm}[]
\caption{Test-Time Contrastive Prompt Guidance for Flow-Based VLA}
\label{alg:cpg}
\begin{algorithmic}[1]
\Require observation $o_k$, novel prompt $\tau^{+}$, trained prompt $\tau^{-}$, guidance scale $w$, step size $\delta<0$
\Ensure final action chunk $a_k^{0}$

\State Sample initial noisy action chunk $a_k^{1} \sim \mathcal{N}(0, I)$
\State Set denoising time $t \gets 1$

\While{$t > 0$}
    \State Forward the same post-trained policy with the novel prompt:
    \State \hspace{1.5em} $v_{\tau^{+}}^{t} \gets v_{\theta}(o_k, \tau^{+}, t)$
    \State Forward the same post-trained policy with the trained prompt:
    \State \hspace{1.5em} $v_{\tau^{-}}^{t} \gets v_{\theta}(o_k, \tau^{-}, t)$
    \State Construct the guided vector field:
    \State \hspace{1.5em} $v_{\mathrm{CPG},k}^{t} \gets v_{\tau^{-}}^{t} + w\!\left(v_{\tau^{+}}^{t} - v_{\tau^{-}}^{t}\right)$
    \State Update the action chunk:
    \State \hspace{1.5em} $a_k^{t+\delta} \gets a_k^{t} + \delta\, v_{\mathrm{CPG},k}^{t}$
    \State $t \gets t + \delta$
\EndWhile

\State \Return $a_k^{0}$
\end{algorithmic}
\end{algorithm}

\section{Post-Training Settings}
\label{app:post_train}

All VLA post-training experiments in this work are built on the official \href{https://github.com/Physical-Intelligence/openpi}{OpenPI} implementation. Unless otherwise specified, all methods are initialized from the same pre-trained model, $\pi_{0.5}\text{-BASE}$~\citep{intelligence2025vision}, to ensure a controlled comparison across post-training strategies.

For {\model} and all of its self-ablations, we use LoRA-based fine-tuning~\citep{hu2022lora}. Specifically, the PaliGemma backbone~\citep{beyer2024paligemma} is frozen, and the inserted LoRA adapters are updated during post-training. Note, however, that following the official OpenPI implementation, the visual encoder is not frozen and remains fully trainable. In contrast, for RETAIN and $\pi_{0.5}\text{-DROID}$, we perform full-parameter fine-tuning. The corresponding model variants and LoRA hyperparameters are summarized in Table~\ref{tab:lora_config}.

\begin{table}[h]
    \centering
    \caption{\textbf{LoRA Model Variants Used During Post-Training.}}
    \label{tab:lora_config}
    \begin{tabular}{lccccccc}
        \toprule
        Module & Width & Depth & MLP Dim & \# Heads & LoRA Rank & LoRA Alpha \\
        \midrule
        Gemma\_2b & 2048 & 18 & 16384 & 8 & 16 & 16.0 \\
        Gemma\_300m & 1024 & 18 & 4096 & 8 & 32 & 32.0 \\
        \bottomrule
    \end{tabular}
\end{table}

In addition, both LoRA-configured modules use a single key-value head (\texttt{num\_kv\_heads}=1) and a head dimension of 256. LoRA adapters are inserted into both the attention layers and feed-forward layers. Specifically, for the Gemma\_2b~(VLM part) we use
\texttt{lora\_configs=\{"attn": rank 16, alpha 16; "ffn": rank 16, alpha 16\}},
and for the Gemma\_300m~(action expert part) we use
\texttt{lora\_configs=\{"attn": rank 32, alpha 32; "ffn": rank 32, alpha 32\}}.

The remaining post-training hyperparameters are shared across experiments unless otherwise noted. We set the action horizon to 10. Training is performed with batch size 32 for a total of 10{,}000 optimization steps for each task. We use AdamW with gradient clipping at a global norm of 1.0. Exponential moving average is disabled.

The learning rate follows a cosine decay schedule with 1{,}000 warmup steps, peak learning rate $5\times 10^{-5}$, and decay steps set to 50{,}000. Since the decay target is also $5\times 10^{-5}$, the schedule effectively maintains a constant learning rate after warmup over our 10{,}000-step training horizon. Table~\ref{tab:posttrain_hparams} summarizes the shared post-training hyperparameters.

\begin{table}[t]
    \centering
    \caption{\textbf{Shared Post-Training Hyperparameters Used Across Methods.}}
    \label{tab:posttrain_hparams}
    \begin{tabular}{ll}
        \toprule
        Hyperparameter & Value \\
        \midrule
        Initialization model & $\pi_{0.5}\text{-BASE}$ \\
        Action horizon & 10 \\
        Discrete state input & False \\
        Batch size & 32 \\
        Optimizer & AdamW \\
        Gradient clipping norm & 1.0 \\
        EMA decay & None \\
        Number of training steps & 10{,}000 \\
        Learning rate schedule & Cosine decay \\
        Warmup steps & 1{,}000 \\
        Peak learning rate & $5\times10^{-5}$ \\
        Decay steps & 50{,}000 \\
        Final decay learning rate & $5\times10^{-5}$ \\
        \bottomrule
    \end{tabular}
\end{table}

\section{Experimental Implementation Details}
\label{app:exp_design}

Our experiments focus on evaluating the generalization capability of post-trained VLA policies. In particular, we study robotic manipulation scenarios where {\model} is post-trained with only a very small amount of task-specific data~(typically around 100 demonstrations for a single task), yet the resulting policy is expected to steer the learned skill toward novel instructions at inference time. 

We evaluate our approach across a diverse set of simulated and real-world manipulation tasks. These include pick-and-place and articulated object interaction, requiring the policy to generalize across diverse object categories and complex spatial targets. These tasks are intentionally constructed so that the post-training data covers only a narrow portion of the instruction and environment configuration space, allowing us to evaluate whether the policy can generalize to novel instructions and configurations. 

All post-training experiments are conducted on NVIDIA A100 GPUs~(80GB). For real-world rollouts, the trained policy is deployed and executed on a workstation equipped with an NVIDIA A5000 GPU.

\subsection{Setup}

\paragraph{Simulation.}
For the simulation experiments, we adopt four environments from the LIBERO benchmark~\citep{liu2023libero}. Because LIBERO was not originally designed to study lock-in effects in low-data VLA post-training, we modify these environments to better expose the failure modes of standard supervised fine-tuning. These adaptations make lock-in behavior more evident, thereby enabling a clearer analysis of the problem setting and a more informative evaluation of our approach.

\begin{figure}[ht]
\centering
\includegraphics[width=0.7\linewidth, trim= 0cm 0cm 0cm 0cm, clip]{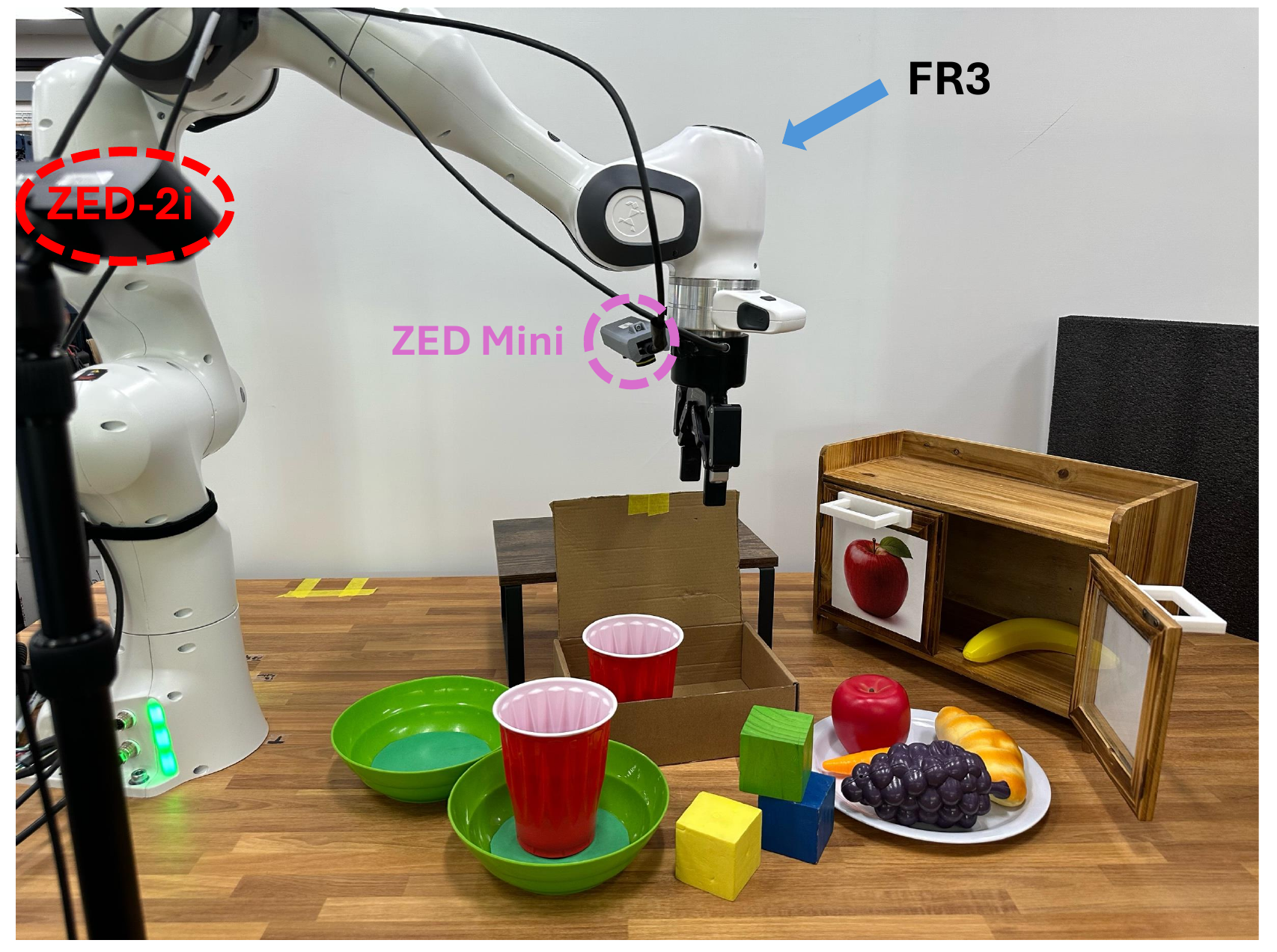}
\caption{\textbf{Real-World Experimental Setup.}}
\label{fig:real_setup}
\end{figure}

\paragraph{Real-World.}
For the real-world experiments, we follow the DROID hardware setup~\citep{khazatsky2024droid}. The robot platform consists of an Franka Research 3~(FR3) arm with a Robotiq 2F-85 gripper. We use two cameras: a third-person ZED-2i camera that captures the robot and the overall scene geometry, and a ZED Mini mounted on the end-effector flange to provide close-up observations for fine-grained manipulation. As in simulation, the real-world tasks are also designed to expose lock-in failure modes in low-data post-training. The real-world setup is shown in Figure~\ref{fig:real_setup}.

\subsection{Task Design}

The benchmark suite is shown in Figure~\ref{fig:benchmark_suite}, with detailed design choice as follows:

\paragraph{MokaPot-on-Stove}
This simulation task is designed to evaluate the \textbf{OOD location-shift} generalization of post-trained VLA policies. In the training demonstrations, the moka pot is randomly initialized within the green-shaded region, and the robot arm reaches for the pot, grasps it by the handle, and places it onto the stove with language instruction {``put moka pot on stove''}. The task is deemed successful if the center of mass of the moka pot lies within a predefined threshold of the stove center. At test time, in addition to evaluating the original demonstration setting, we also initialize the moka pot randomly in the blue-shaded region, which is completely out of distribution with respect to object location, and evaluate whether the post-trained policy can still complete the task successfully.

\paragraph{Mug-on-Plate~[C]}
This simulation task is designed to evaluate the \textbf{concept lock-in} behavior of post-trained VLA policies. In the training demonstrations, multiple mugs are randomly placed on the table, and the robot is instructed to pick and place a specific-colored mug onto a plate. During post-training, the target object is always the green mug, and the training prompt is fixed to {``put green mug on plate''}. The robot reaches the target mug, grasps it, and places it onto the plate. The task is considered successful when the center of mass of the mug lies within a predefined threshold of the plate center. At test time, in addition to evaluating the original demonstration setting, we replace the target instruction with novel color concepts that are never used as targets during post-training (e.g., {``put red mug on plate''} or {``put blue mug on plate''}) and evaluate whether the post-trained policy can follow these instructions correctly. Because the green mug is observed at diverse table locations during training, this task largely factors out location generalization and isolates failure caused by overfitting to the seen object concept.

\paragraph{Block-Stacking~[C]}
This real-world task is also designed to evaluate the \textbf{concept lock-in} behavior of post-trained VLA policies. Similar to \textsc{Mug-on-Plate~[C]}, the goal is to test whether a policy post-trained on a single seen concept can still follow novel object-specific instructions at test time. In the training demonstrations, multiple colored blocks are randomly placed on the table, and the robot is instructed to pick one specific block and stack it on top of another fixed target block. During post-training, the prompt is fixed to a single seen object concept, e.g., {``stack blue block on green block''}. The robot reaches the target block, grasps it, and places it onto the target support block. The task is considered successful when the manipulated block is stably positioned on top of the target block within a predefined spatial threshold. At test time, in addition to evaluating the original demonstration setting, we also prompt the policy with unseen object instructions that refer to different source blocks. This tests whether the post-trained policy can correctly follow novel concept-level instructions rather than overfitting to the seen object identity used during training.

\paragraph{Food-on-Plate~[C]}
This real-world task likewise evaluates \textbf{concept lock-in}, with the same motivation as in \textsc{Mug-on-Plate~[C]}. In the training demonstrations, multiple food objects are randomly placed on the table, and the robot is instructed to pick one specific food item and place it onto a plate. During post-training, the target object is always a single seen concept, with the prompt fixed to {``put bread on plate''}. The robot reaches the bread, grasps it, and places it onto the plate. The task is considered successful when the center of mass of the food object lies within a predefined threshold of the plate center. At test time, besides evaluating the original demonstration setting, we also replace the target instruction with novel food concepts that are never used as targets during post-training, such as {``put apple on plate''} or {``put carrot on plate''}. As in \textsc{Mug-on-Plate [C]}, the seen training object appears at diverse locations across demonstrations, so failure in this task primarily reflects overfitting to the seen concept rather than an inability to reach different table locations. In addition, we also evaluate \textbf{OOD location-shift} generalization in this task. During training, the plate is randomly initialized within the green-shaded region, whereas at test time we additionally initialize it within the blue-shaded region.

\paragraph{Open-Microwave~[S]}
This simulation task is designed to evaluate the \textbf{spatial lock-in} behavior of post-trained VLA policies. Unlike the concept-generalization tasks above, the novel test instructions here do not introduce new object concepts; instead, they require the policy to respond to spatial information expressed in the prompt. The environment contains two microwaves arranged vertically, referred to as the lower microwave and the upper microwave. In the training demonstrations, the robot is instructed to open only the lower microwave, with the prompt fixed to {``open lower microwave''}. The robot reaches the handle of the lower microwave and executes the corresponding opening motion. The task is considered successful when the target microwave door is opened beyond a predefined angle threshold. At test time, in addition to evaluating the original demonstration setting, we also evaluate the policy under the novel spatial instruction {``open upper microwave''}. Since the instruction introduces no new object concept and only changes the spatial referent, this task specifically tests whether the post-trained policy can overcome spatial lock-in and generalize the learned manipulation behavior to a previously uncovered region of the workspace. The observations in this task are provided using only wrist-view image.

\paragraph{Mug-on-Plate~[S]}
This simulation task is designed to evaluate the \textbf{spatial lock-in} behavior of post-trained VLA policies. The object concept remains unchanged throughout training and testing; the challenge is instead whether the policy can follow a novel spatial instruction and move the object to a target region not covered during post-training. In the training demonstrations, the robot is always instructed to place the mug onto one specific plate, with the prompt fixed to {``put mug on left plate''}. The robot reaches the mug, grasps it, and places it onto the designated plate. The task is considered successful when the center of mass of the mug lies within a predefined threshold of the target plate center. At test time, in addition to evaluating the original demonstration setting, we also evaluate a novel spatial instruction, {``put mug on right plate''}. Since the instruction introduces no new concept and only changes the spatial target, this task specifically tests whether the post-trained policy can overcome spatial lock-in and generalize the learned manipulation skill to a previously uncovered target region. The observations in this task are provided using only wrist-view image. The left/right concept is defined in wrist-view camera.

\paragraph{Cup-to-Box~[S]}
This real-world task also evaluates \textbf{spatial lock-in}, following the same rationale as in \textsc{Mug-on-Plate~[S]}. In the training demonstrations, the robot is instructed to move a specific cup into a box, with the prompt fixed to {``put right cup to box''}. The robot reaches the designated cup, grasps it, and places it into the box. The task is considered successful when the targeted cup is placed inside the box. At test time, besides evaluating the original demonstration setting, we also prompt the policy with a novel spatial instruction, {``put left cup to box''}, which requires the policy to manipulate the same object concept but in a target region not covered during training. This task therefore isolates whether the post-trained policy can use prompt-level spatial information to guide manipulation beyond the narrow spatial support of the demonstrations. The left/right concept is defined in third-view camera.

\paragraph{Open-Labeled-Door~[C+S]}
This real-world task is designed to evaluate the combined \textbf{concept and spatial lock-in} behavior of post-trained VLA policies. The environment contains a cabinet with two doors that have a fixed spatial relationship: one door is on the left and the other is on the right. To distinguish the two doors, we use visual labels attached to them, e.g., banana and apple images. In the training demonstrations, the robot is instructed to open only one labeled door, with the prompt fixed to a single seen instruction such as {``open door labeled with banana''}. The robot reaches the corresponding handle and pulls the door open. The task is considered successful when the specified door is opened beyond a predefined threshold. At test time, in addition to evaluating the original demonstration setting, we also prompt the policy to open the other labeled door, e.g., {``open door labeled with apple''}. This requires the policy to overcome \emph{spatial lock-in}, since the unseen target door occupies a different spatial region that is never opened during training, and also to overcome \emph{concept lock-in}, since the two doors are referred to through different visual labels and the novel instruction changes which label should be grounded to action.

\section{Additional Experimental Results}
\label{app:exp_result}

\subsection{In-Distribution Evaluation Results}

Table~\ref{tab:quant_id} reports the in-distribution performance of {\model} and all baselines under the trained prompts. The results show that all methods perform well in-distribution after low-data post-training, indicating that the main challenge is not fitting the demonstrated skill itself, but generalizing beyond the post-training instruction coverage.

\begin{table}[]
\caption{\textbf{In-distribution Success Counts~(20 trials per task).}}
\label{tab:quant_id}
\centering
\footnotesize
\setlength{\tabcolsep}{4pt}
\renewcommand{\arraystretch}{1.0}
\hspace*{-0.02\textwidth}%
\begin{minipage}{1.04\textwidth}
\centering
\makebox[\textwidth][c]{%
\begin{tabular}{@{}lccccccc@{}}
\toprule
Method & T2 [C] & T3 [C] & T4 [C] & T5 [S] & T6 [S] & T7 [S] & T8 [C+S] \\
\midrule
RETAIN & \textbf{20}/20 & 17/20 & \textbf{20}/20 & \textbf{20}/20 & \textbf{20}/20 & 18/20 & 16/20 \\
$\pi_{0.5}\text{-DROID}$ & -- & 18/20 & 17/20 & -- & -- & 8/20 & 0/20 \\
{\model} w/o CPG & \textbf{20}/20 & \textbf{19}/20 & \textbf{20}/20 & \textbf{20}/20 & \textbf{20}/20 & 19/20 & \textbf{19}/20 \\
{\model} w/o Vis-Reg & 16/20 & 18/20 & 19/20 & \textbf{20}/20 & \textbf{20}/20 & \textbf{20}/20 & \textbf{19}/20 \\
{\model} w/ Frozen-Vis & 18/20 & 18/20 & 15/20 & 19/20 & \textbf{20}/20 & 17/20 & 13/20 \\
{\model} & \textbf{20}/20 & 18/20 & \textbf{20}/20 & \textbf{20}/20 & \textbf{20}/20 & \textbf{20}/20 & 18/20 \\
\bottomrule
\end{tabular}%
}
\end{minipage}
\end{table}

\subsection{Full Qualitative Novel-Prompt Rollout Results}

We present full qualitative rollout results of {\model} under OOD settings. Notably, in \textsc{Food-on-Plate}~[C], we introduce banana and grapes into the scene, although \textbf{neither object appears} during the SFT stage. We then prompt the policy to ``grasp banana'' and place it onto the OOD target plate. Despite this combined concept and location shift, the policy is able to follow the novel instruction and successfully complete the task.

\begin{figure}[ht]
\centering
\includegraphics[width=0.99\linewidth, trim= 0cm 0cm 0cm 0cm, clip]{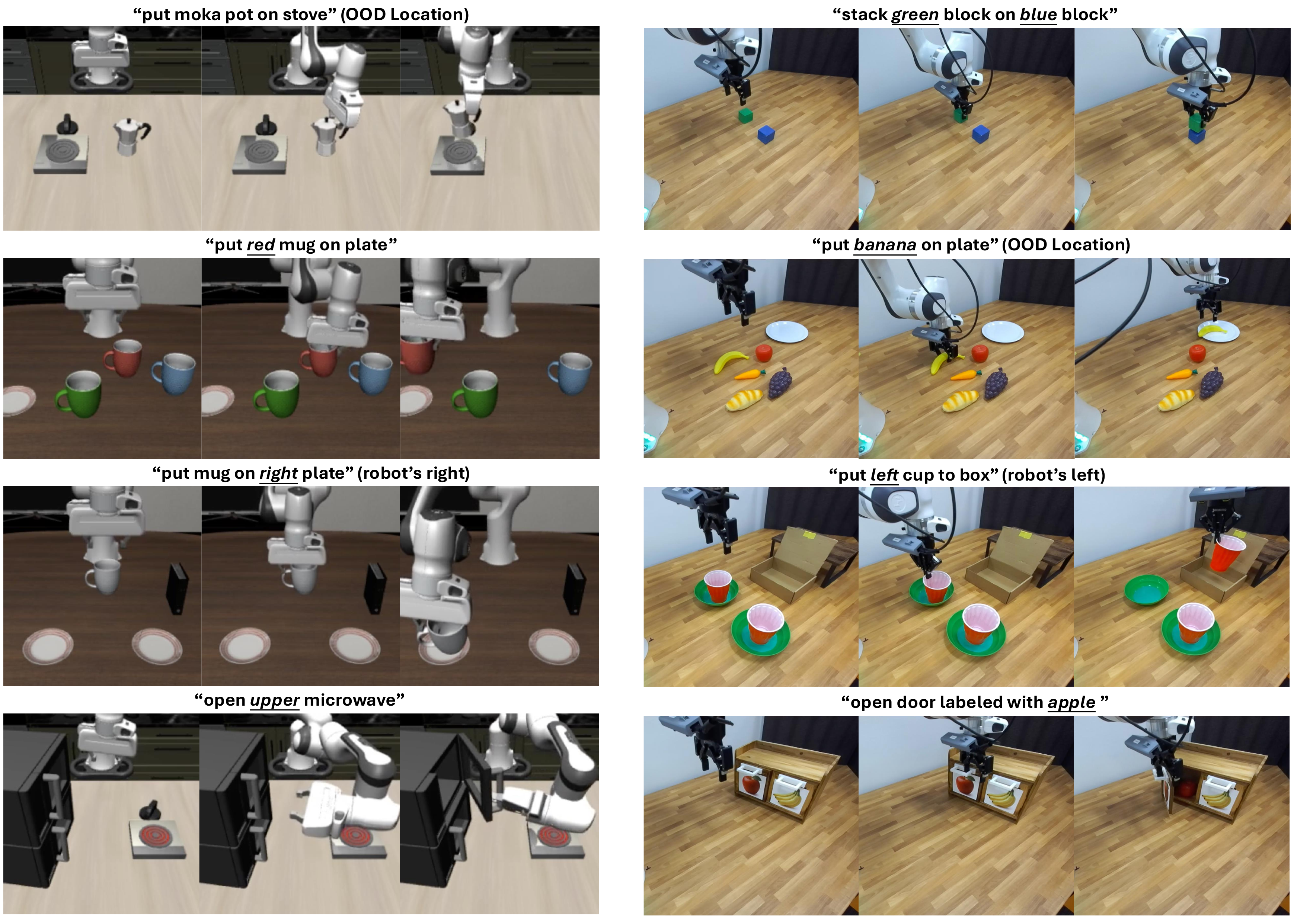}
\caption{\textbf{Full Qualitative Results with Novel Prompts.}}
\label{fig:full_qual_results}
\end{figure}

Additionally, in the \textsc{Mug-on-Plate}~[S] task, we place the book at the center of the scene and directly prompt the policy with ``put book on left plate'' without any further fine-tuning. {\model} is able to reach the book and place it on the left plate, whereas the standard fine-tuning baseline produces a largely random trajectory. This result suggests that {\model} can transfer the learned spatial behavior to a novel object specified only through the prompt, without any additional fine-tuning.

\begin{figure}[ht]
\centering
\includegraphics[width=0.8\linewidth, trim= 0cm 0cm 0cm 0cm, clip]{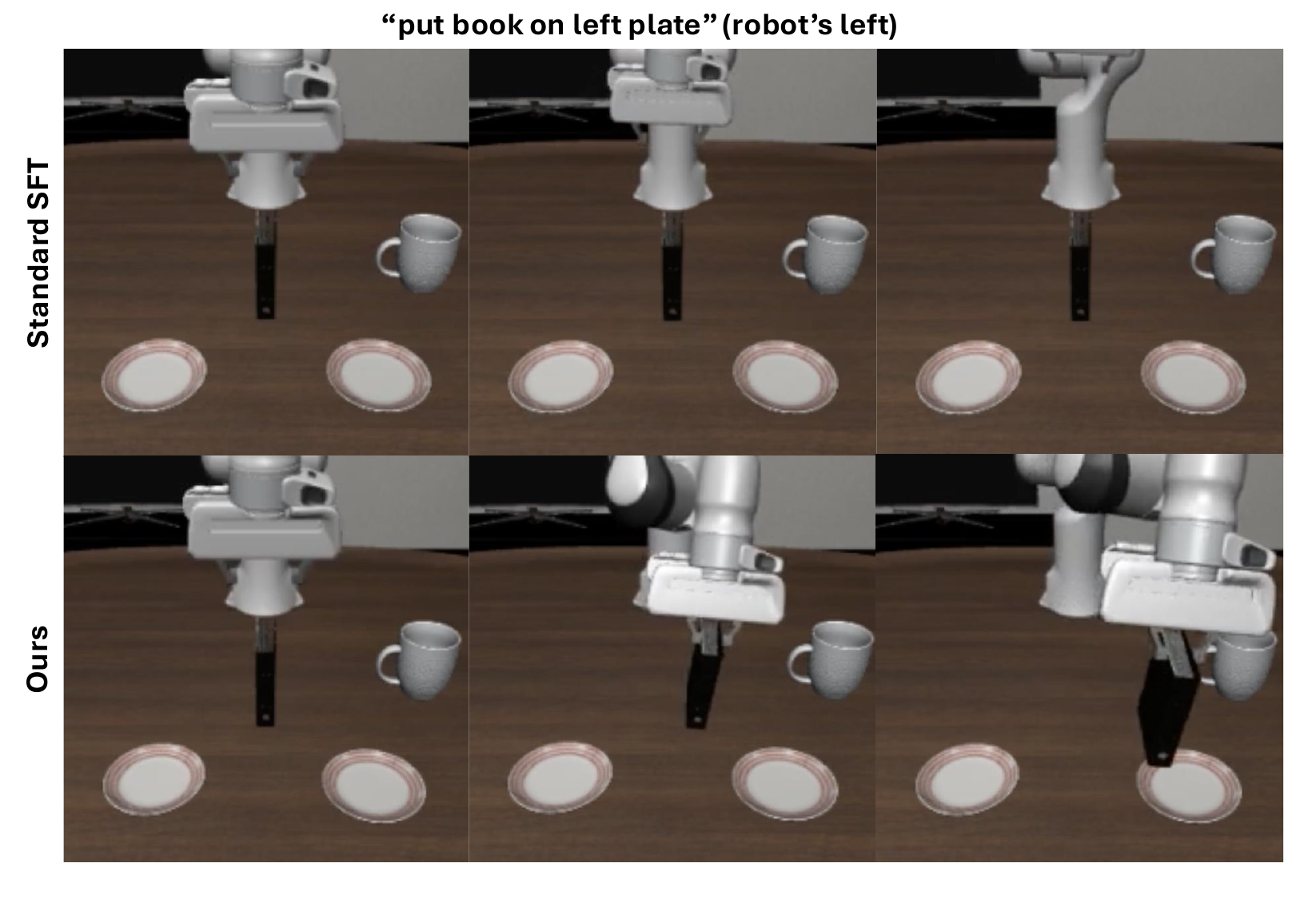}
\caption{\textbf{Prompted for a Novel Object.}}
\label{fig:black_book}
\end{figure}

\subsection{CPG with an Invalid Positive Prompt}

Finally, we test whether CPG succeeds specifically by leveraging the semantic content of the positive prompt. We again consider the \textsc{Mug-on-Plate}~[S] task, but replace the intended positive prompt $\tau^{+}$ with the malformed instruction ``put mug on write plate''. Since ``write'' does not specify a meaningful spatial target, the resulting positive condition no longer provides the information needed to steer the policy toward the right plate. In this case, the robot reverts to the post-training bias and places the mug on the left plate. This result suggests that CPG is effective only when the positive prompt provides valid task-relevant semantics; when it succeeds, the resulting behavior is specifically steered by the information encoded in the positive prompt.

\begin{figure}[ht]
\centering
\includegraphics[width=0.8\linewidth, trim= 0cm 0cm 0cm 0cm, clip]{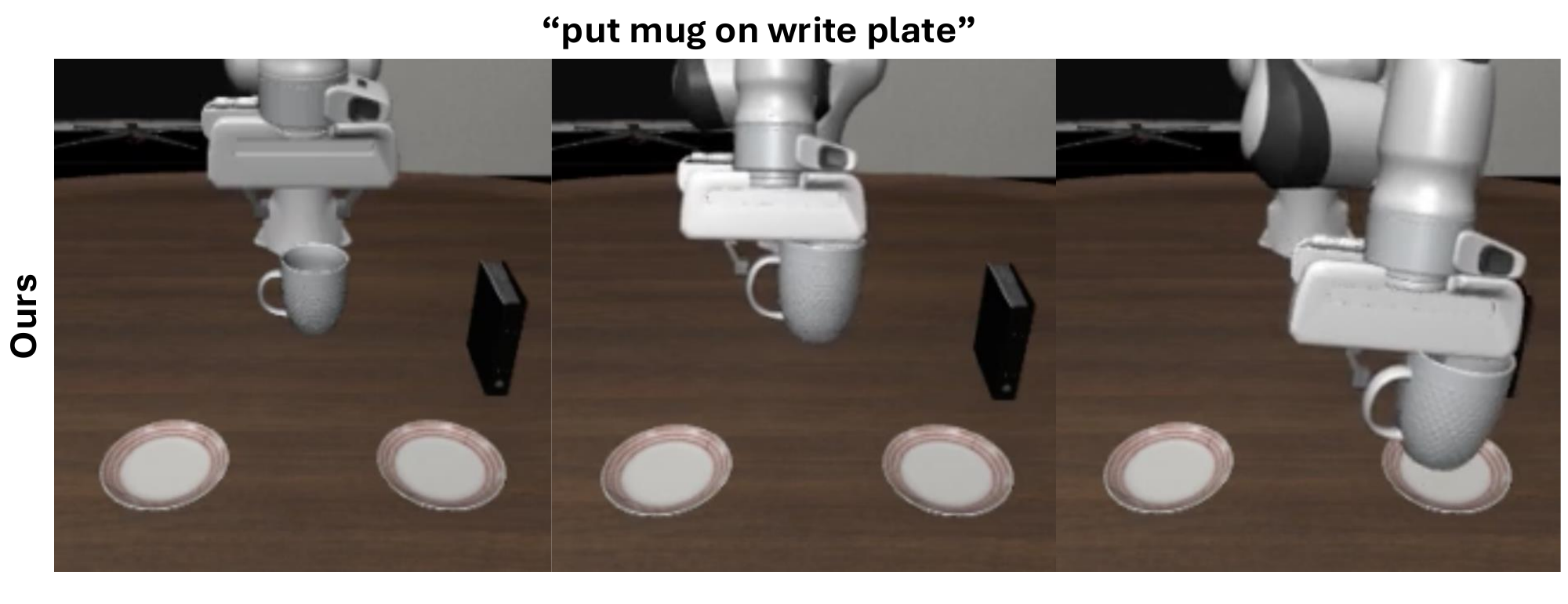}
\caption{\textbf{CPG with an Invalid Positive Prompt.}}
\label{fig:wrong_prompt}
\end{figure}

\subsection{Comparison with a Baseline that Uses Foundation-Model Supervision During Post-Training}

We also compare {\model} against a strong baseline, Spatial Forcing~\citep{li2025spatial}, which incorporates additional supervision from an external 3D foundation model during post-training to encourage implicit spatial representation alignment. We evaluate both methods on the four simulation tasks in our lock-in benchmark. As shown in Table~\ref{tab:quant_sf}, despite leveraging additional external supervision, Spatial Forcing exhibits very limited adaptation to novel situations across all four tasks, whereas {\model} achieves substantially stronger generalization.

\begin{table}[t]
\caption{\textbf{Generalization Comparison in Simulation Tasks~(20 trials per task).}}
\label{tab:quant_sf}
\centering
\footnotesize
\setlength{\tabcolsep}{4pt}
\renewcommand{\arraystretch}{1.0}
\hspace*{-0.02\textwidth}%
\begin{minipage}{1.04\textwidth}
\centering
\makebox[\textwidth][c]{%
\begin{tabular}{@{}lcccc@{}}
\toprule
Method & T1 & T2 [C] & T5 [S] & T6 [S] \\
\midrule
Spatial Forcing & 2/20 & 1/20 & 0/20 & 0/20 \\
{\model} & \textbf{16}/20 & \textbf{19}/20 & \textbf{11}/20 & \textbf{13}/20 \\
\bottomrule
\end{tabular}%
}
\end{minipage}
\end{table}

%% file: main.bib
@article{intelligence2025vision,
  title={pi\_05: a vision-language-action model with open-world generalization},
  author={Intelligence, Physical and Black, Kevin and Brown, Noah and Darpinian, James and Dhabalia, Karan and Driess, Danny and Esmail, Adnan and Equi, Michael and Finn, Chelsea and Fusai, Niccolo and others},
  journal={pi05: a vision-language-action model with open-world generalization},
  year={2025}
}

@article{khazatsky2024droid,
  title={Droid: A large-scale in-the-wild robot manipulation dataset},
  author={Khazatsky, Alexander and Pertsch, Karl and Nair, Suraj and Balakrishna, Ashwin and Dasari, Sudeep and Karamcheti, Siddharth and Nasiriany, Soroush and Srirama, Mohan Kumar and Chen, Lawrence Yunliang and Ellis, Kirsty and others},
  journal={arXiv preprint arXiv:2403.12945},
  year={2024}
}

@article{kim2024openvla,
  title={Openvla: An open-source vision-language-action model},
  author={Kim, Moo Jin and Pertsch, Karl and Karamcheti, Siddharth and Xiao, Ted and Balakrishna, Ashwin and Nair, Suraj and Rafailov, Rafael and Foster, Ethan and Lam, Grace and Sanketi, Pannag and others},
  journal={arXiv preprint arXiv:2406.09246},
  year={2024}
}

@article{bjorck2025gr00t,
  title={Gr00t n1: An open foundation model for generalist humanoid robots},
  author={Bjorck, Johan and Casta{\~n}eda, Fernando and Cherniadev, Nikita and Da, Xingye and Ding, Runyu and Fan, Linxi and Fang, Yu and Fox, Dieter and Hu, Fengyuan and Huang, Spencer and others},
  journal={arXiv preprint arXiv:2503.14734},
  year={2025}
}

@article{pertsch2025fast,
  title={Fast: Efficient action tokenization for vision-language-action models},
  author={Pertsch, Karl and Stachowicz, Kyle and Ichter, Brian and Driess, Danny and Nair, Suraj and Vuong, Quan and Mees, Oier and Finn, Chelsea and Levine, Sergey},
  journal={arXiv preprint arXiv:2501.09747},
  year={2025}
}

@article{guo2025ctrl,
  title={Ctrl-world: A controllable generative world model for robot manipulation},
  author={Guo, Yanjiang and Shi, Lucy Xiaoyang and Chen, Jianyu and Finn, Chelsea},
  journal={arXiv preprint arXiv:2510.10125},
  year={2025}
}

@article{guo2026vlaw,
  title={VLAW: Iterative Co-Improvement of Vision-Language-Action Policy and World Model},
  author={Guo, Yanjiang and Lee, Tony and Shi, Lucy Xiaoyang and Chen, Jianyu and Liang, Percy and Finn, Chelsea},
  journal={arXiv preprint arXiv:2602.12063},
  year={2026}
}

@inproceedings{o2024open,
  title={Open x-embodiment: Robotic learning datasets and rt-x models: Open x-embodiment collaboration 0},
  author={O’Neill, Abby and Rehman, Abdul and Maddukuri, Abhiram and Gupta, Abhishek and Padalkar, Abhishek and Lee, Abraham and Pooley, Acorn and Gupta, Agrim and Mandlekar, Ajay and Jain, Ajinkya and others},
  booktitle={2024 IEEE International Conference on Robotics and Automation (ICRA)},
  pages={6892--6903},
  year={2024},
  organization={IEEE}
}

@article{bousmalis2023robocat,
  title={Robocat: A self-improving generalist agent for robotic manipulation},
  author={Bousmalis, Konstantinos and Vezzani, Giulia and Rao, Dushyant and Devin, Coline and Lee, Alex X and Bauz{\'a}, Maria and Davchev, Todor and Zhou, Yuxiang and Gupta, Agrim and Raju, Akhil and others},
  journal={arXiv preprint arXiv:2306.11706},
  year={2023}
}

@article{black2024pi_0,
  title={pi\_0: A Vision-Language-Action Flow Model for General Robot Control},
  author={Black, Kevin and Brown, Noah and Driess, Danny and Esmail, Adnan and Equi, Michael and Finn, Chelsea and Fusai, Niccolo and Groom, Lachy and Hausman, Karol and Ichter, Brian and others},
  journal={arXiv preprint arXiv:2410.24164},
  year={2024}
}

@article{li2025controlvla,
  title={Controlvla: Few-shot object-centric adaptation for pre-trained vision-language-action models},
  author={Li, Puhao and Wu, Yingying and Xi, Ziheng and Li, Wanlin and Huang, Yuzhe and Zhang, Zhiyuan and Chen, Yinghan and Wang, Jianan and Zhu, Song-Chun and Liu, Tengyu and others},
  journal={arXiv preprint arXiv:2506.16211},
  year={2025}
}

@article{team2025gemini,
  title={Gemini robotics: Bringing ai into the physical world},
  author={Team, Gemini Robotics and Abeyruwan, Saminda and Ainslie, Joshua and Alayrac, Jean-Baptiste and Arenas, Montserrat Gonzalez and Armstrong, Travis and Balakrishna, Ashwin and Baruch, Robert and Bauza, Maria and Blokzijl, Michiel and others},
  journal={arXiv preprint arXiv:2503.20020},
  year={2025}
}

@article{cheng2025moe,
  title={MoE-DP: An MoE-Enhanced Diffusion Policy for Robust Long-Horizon Robotic Manipulation with Skill Decomposition and Failure Recovery},
  author={Cheng, Baiye and Liang, Tianhai and Huang, Suning and Shao, Maanping and Zhang, Feihong and Xu, Botian and Xue, Zhengrong and Xu, Huazhe},
  journal={arXiv preprint arXiv:2511.05007},
  year={2025}
}

@inproceedings{guo2025improving,
  title={Improving vision-language-action model with online reinforcement learning},
  author={Guo, Yanjiang and Zhang, Jianke and Chen, Xiaoyu and Ji, Xiang and Wang, Yen-Jen and Hu, Yucheng and Chen, Jianyu},
  booktitle={2025 IEEE International Conference on Robotics and Automation (ICRA)},
  pages={15665--15672},
  year={2025},
  organization={IEEE}
}

@inproceedings{wortsman2022robust,
  title={Robust fine-tuning of zero-shot models},
  author={Wortsman, Mitchell and Ilharco, Gabriel and Kim, Jong Wook and Li, Mike and Kornblith, Simon and Roelofs, Rebecca and Lopes, Raphael Gontijo and Hajishirzi, Hannaneh and Farhadi, Ali and Namkoong, Hongseok and others},
  booktitle={Proceedings of the IEEE/CVF conference on computer vision and pattern recognition},
  pages={7959--7971},
  year={2022}
}

@article{jin2022dataless,
  title={Dataless knowledge fusion by merging weights of language models},
  author={Jin, Xisen and Ren, Xiang and Preotiuc-Pietro, Daniel and Cheng, Pengxiang},
  journal={arXiv preprint arXiv:2212.09849},
  year={2022}
}

@article{zhou2025libero,
  title={LIBERO-PRO: Towards Robust and Fair Evaluation of Vision-Language-Action Models Beyond Memorization},
  author={Zhou, Xueyang and Xu, Yangming and Tie, Guiyao and Chen, Yongchao and Zhang, Guowen and Chu, Duanfeng and Zhou, Pan and Sun, Lichao},
  journal={arXiv preprint arXiv:2510.03827},
  year={2025}
}

@article{fei2025libero,
  title={Libero-plus: In-depth robustness analysis of vision-language-action models},
  author={Fei, Senyu and Wang, Siyin and Shi, Junhao and Dai, Zihao and Cai, Jikun and Qian, Pengfang and Ji, Li and He, Xinzhe and Zhang, Shiduo and Fei, Zhaoye and others},
  journal={arXiv preprint arXiv:2510.13626},
  year={2025}
}

@article{zang2025rlinf,
  title={Rlinf-vla: A unified and efficient framework for vla+ rl training},
  author={Zang, Hongzhi and Wei, Mingjie and Xu, Si and Wu, Yongji and Guo, Zhen and Wang, Yuanqing and Lin, Hao and Shi, Liangzhi and Xie, Yuqing and Xu, Zhexuan and others},
  journal={arXiv preprint arXiv:2510.06710},
  year={2025}
}

@article{liu2026vls,
  title={VLS: Steering Pretrained Robot Policies via Vision-Language Models},
  author={Liu, Shuo and Singh, Ishneet Sukhvinder and Xu, Yiqing and Duan, Jiafei and Krishna, Ranjay},
  journal={arXiv preprint arXiv:2602.03973},
  year={2026}
}

@article{chen2026steerable,
  title={Steerable Vision-Language-Action Policies for Embodied Reasoning and Hierarchical Control},
  author={Chen, William and Bhatia, Jagdeep Singh and Glossop, Catherine and Mathihalli, Nikhil and Doshi, Ria and Tang, Andy and Driess, Danny and Pertsch, Karl and Levine, Sergey},
  journal={arXiv preprint arXiv:2602.13193},
  year={2026}
}

@article{li2025spatial,
  title={Spatial forcing: Implicit spatial representation alignment for vision-language-action model},
  author={Li, Fuhao and Song, Wenxuan and Zhao, Han and Wang, Jingbo and Ding, Pengxiang and Wang, Donglin and Zeng, Long and Li, Haoang},
  journal={arXiv preprint arXiv:2510.12276},
  year={2025}
}

@article{grover2025enhancing,
  title={Enhancing generalization in vision-language-action models by preserving pretrained representations},
  author={Grover, Shresth and Gopalkrishnan, Akshay and Ai, Bo and Christensen, Henrik I and Su, Hao and Li, Xuanlin},
  journal={arXiv preprint arXiv:2509.11417},
  year={2025}
}

@article{kachaev2510don,
  title={Don’t blind your vla: Aligning visual representations for ood generalization, 2025},
  author={Kachaev, Nikita and Kolosov, Mikhail and Zelezetsky, Daniil and Kovalev, Alexey K and Panov, Aleksandr I},
  journal={URL https://arxiv. org/abs/2510.25616},
  volume={2},
  number={4}
}

@article{lepert2025masquerade,
  title={Masquerade: Learning from in-the-wild human videos using data-editing},
  author={Lepert, Marion and Fang, Jiaying and Bohg, Jeannette},
  journal={arXiv preprint arXiv:2508.09976},
  year={2025}
}

@inproceedings{punamiya2025egobridge,
  title={Egobridge: Domain adaptation for generalizable imitation from egocentric human data},
  author={Punamiya, Ryan and Patel, Dhruv and Aphiwetsa, Patcharapong and Kuppili, Pranav and Zhu, Lawrence Y and Kareer, Simar and Hoffman, Judy and Xu, Danfei},
  booktitle={Human to Robot: Workshop on Sensorizing, Modeling, and Learning from Humans},
  year={2025}
}

@article{yadav2025robust,
  title={Robust Finetuning of Vision-Language-Action Robot Policies via Parameter Merging},
  author={Yadav, Yajat and Zhou, Zhiyuan and Wagenmaker, Andrew and Pertsch, Karl and Levine, Sergey},
  journal={arXiv preprint arXiv:2512.08333},
  year={2025}
}

@article{ma2024survey,
  title={A survey on vision-language-action models for embodied ai},
  author={Ma, Yueen and Song, Zixing and Zhuang, Yuzheng and Hao, Jianye and King, Irwin},
  journal={arXiv preprint arXiv:2405.14093},
  year={2024}
}

@article{bommasani2021opportunities,
  title={On the opportunities and risks of foundation models},
  author={Bommasani, Rishi and Hudson, Drew A and Adeli, Ehsan and Altman, Russ and Arora, Simran and von Arx, Sydney and Bernstein, Michael S and Bohg, Jeannette and Bosselut, Antoine and Brunskill, Emma and others},
  journal={arXiv preprint arXiv:2108.07258},
  year={2021}
}

@article{koulischer2024dynamic,
  title={Dynamic negative guidance of diffusion models},
  author={Koulischer, Felix and Deleu, Johannes and Raya, Gabriel and Demeester, Thomas and Ambrogioni, Luca},
  journal={arXiv preprint arXiv:2410.14398},
  year={2024}
}

@inproceedings{nguyen2024language,
  title={Language-driven 6-dof grasp detection using negative prompt guidance},
  author={Nguyen, Toan and Vu, Minh Nhat and Huang, Baoru and Vuong, An and Vuong, Quan and Le, Ngan and Vo, Thieu and Nguyen, Anh},
  booktitle={European Conference on Computer Vision},
  pages={363--381},
  year={2024},
  organization={Springer}
}

@inproceedings{ban2024understanding,
  title={Understanding the impact of negative prompts: When and how do they take effect?},
  author={Ban, Yuanhao and Wang, Ruochen and Zhou, Tianyi and Cheng, Minhao and Gong, Boqing and Hsieh, Cho-Jui},
  booktitle={european conference on computer vision},
  pages={190--206},
  year={2024},
  organization={Springer}
}

@inproceedings{jang2023can,
  title={Can large language models truly understand prompts? a case study with negated prompts},
  author={Jang, Joel and Ye, Seonghyeon and Seo, Minjoon},
  booktitle={Transfer learning for natural language processing workshop},
  pages={52--62},
  year={2023},
  organization={PMLR}
}

@article{liu2023libero,
  title={Libero: Benchmarking knowledge transfer for lifelong robot learning},
  author={Liu, Bo and Zhu, Yifeng and Gao, Chongkai and Feng, Yihao and Liu, Qiang and Zhu, Yuke and Stone, Peter},
  journal={Advances in Neural Information Processing Systems},
  volume={36},
  pages={44776--44791},
  year={2023}
}

@article{shenfeld2025rl,
  title={Rl's razor: Why online reinforcement learning forgets less},
  author={Shenfeld, Idan and Pari, Jyothish and Agrawal, Pulkit},
  journal={arXiv preprint arXiv:2509.04259},
  year={2025}
}

@article{du2025dynaguide,
  title={Dynaguide: Steering diffusion polices with active dynamic guidance},
  author={Du, Maximilian and Song, Shuran},
  journal={arXiv preprint arXiv:2506.13922},
  year={2025}
}

@inproceedings{zitkovich2023rt,
  title={Rt-2: Vision-language-action models transfer web knowledge to robotic control},
  author={Zitkovich, Brianna and Yu, Tianhe and Xu, Sichun and Xu, Peng and Xiao, Ted and Xia, Fei and Wu, Jialin and Wohlhart, Paul and Welker, Stefan and Wahid, Ayzaan and others},
  booktitle={Conference on Robot Learning},
  pages={2165--2183},
  year={2023},
  organization={PMLR}
}

@article{anil2023palm,
  title={Palm 2 technical report},
  author={Anil, Rohan and Dai, Andrew M and Firat, Orhan and Johnson, Melvin and Lepikhin, Dmitry and Passos, Alexandre and Shakeri, Siamak and Taropa, Emanuel and Bailey, Paige and Chen, Zhifeng and others},
  journal={arXiv preprint arXiv:2305.10403},
  year={2023}
}

@article{black2023zero,
  title={Zero-shot robotic manipulation with pretrained image-editing diffusion models},
  author={Black, Kevin and Nakamoto, Mitsuhiko and Atreya, Pranav and Walke, Homer and Finn, Chelsea and Kumar, Aviral and Levine, Sergey},
  journal={arXiv preprint arXiv:2310.10639},
  year={2023}
}

@article{team2024octo,
  title={Octo: An open-source generalist robot policy},
  author={Team, Octo Model and Ghosh, Dibya and Walke, Homer and Pertsch, Karl and Black, Kevin and Mees, Oier and Dasari, Sudeep and Hejna, Joey and Kreiman, Tobias and Xu, Charles and others},
  journal={arXiv preprint arXiv:2405.12213},
  year={2024}
}

@article{sharma2023lossless,
  title={Lossless adaptation of pretrained vision models for robotic manipulation},
  author={Sharma, Mohit and Fantacci, Claudio and Zhou, Yuxiang and Koppula, Skanda and Heess, Nicolas and Scholz, Jon and Aytar, Yusuf},
  journal={arXiv preprint arXiv:2304.06600},
  year={2023}
}

@article{fu2024context,
  title={In-context imitation learning via next-token prediction},
  author={Fu, Letian and Huang, Huang and Datta, Gaurav and Chen, Lawrence Yunliang and Panitch, William Chung-Ho and Liu, Fangchen and Li, Hui and Goldberg, Ken},
  journal={arXiv preprint arXiv:2408.15980},
  year={2024}
}

@article{huang2024mentor,
  title={Mentor: Mixture-of-experts network with task-oriented perturbation for visual reinforcement learning},
  author={Huang, Suning and Zhang, Zheyu and Liang, Tianhai and Xu, Yihan and Kou, Zhehao and Lu, Chenhao and Xu, Guowei and Xue, Zhengrong and Xu, Huazhe},
  journal={arXiv preprint arXiv:2410.14972},
  year={2024}
}

@article{chen2025sarm,
  title={SARM: Stage-Aware Reward Modeling for Long Horizon Robot Manipulation},
  author={Chen, Qianzhong and Yu, Justin and Schwager, Mac and Abbeel, Pieter and Shentu, Yide and Wu, Philipp},
  journal={arXiv preprint arXiv:2509.25358},
  year={2025}
}

@article{li2025simplevla,
  title={Simplevla-rl: Scaling vla training via reinforcement learning},
  author={Li, Haozhan and Zuo, Yuxin and Yu, Jiale and Zhang, Yuhao and Yang, Zhaohui and Zhang, Kaiyan and Zhu, Xuekai and Zhang, Yuchen and Chen, Tianxing and Cui, Ganqu and others},
  journal={arXiv preprint arXiv:2509.09674},
  year={2025}
}

@article{pan2026sop,
  title={SOP: A Scalable Online Post-Training System for Vision-Language-Action Models},
  author={Pan, Mingjie and Feng, Siyuan and Zhang, Qinglin and Li, Xinchen and Song, Jianheng and Qu, Chendi and Wang, Yi and Li, Chuankang and Xiong, Ziyu and Chen, Zhi and others},
  journal={arXiv preprint arXiv:2601.03044},
  year={2026}
}

@article{xing2025shortcut,
  title={Shortcut learning in generalist robot policies: The role of dataset diversity and fragmentation},
  author={Xing, Youguang and Luo, Xu and Xie, Junlin and Gao, Lianli and Shen, Hengtao and Song, Jingkuan},
  journal={arXiv preprint arXiv:2508.06426},
  year={2025}
}

@article{xiang2025parallels,
  title={Parallels between vla model post-training and human motor learning: Progress, challenges, and trends},
  author={Xiang, Tian-Yu and Jin, Ao-Qun and Zhou, Xiao-Hu and Gui, Mei-Jiang and Xie, Xiao-Liang and Liu, Shi-Qi and Wang, Shuang-Yi and Duan, Sheng-Bin and Xie, Fu-Chao and Wang, Wen-Kai and others},
  journal={arXiv preprint arXiv:2506.20966},
  year={2025}
}

@article{hu2022lora,
  title={Lora: Low-rank adaptation of large language models.},
  author={Hu, Edward J and Shen, Yelong and Wallis, Phillip and Allen-Zhu, Zeyuan and Li, Yuanzhi and Wang, Shean and Wang, Liang and Chen, Weizhu and others},
  journal={Iclr},
  volume={1},
  number={2},
  pages={3},
  year={2022}
}

@article{sun2025latent,
  title={Latent policy barrier: Learning robust visuomotor policies by staying in-distribution},
  author={Sun, Zhanyi and Song, Shuran},
  journal={arXiv preprint arXiv:2508.05941},
  year={2025}
}

@article{huang2025particleformer,
  title={Particleformer: A 3d point cloud world model for multi-object, multi-material robotic manipulation},
  author={Huang, Suning and Chen, Qianzhong and Zhang, Xiaohan and Sun, Jiankai and Schwager, Mac},
  journal={arXiv preprint arXiv:2506.23126},
  year={2025}
}

@article{nakamoto2024steering,
  title={Steering your generalists: Improving robotic foundation models via value guidance},
  author={Nakamoto, Mitsuhiko and Mees, Oier and Kumar, Aviral and Levine, Sergey},
  journal={arXiv preprint arXiv:2410.13816},
  year={2024}
}

@article{cao2025compose,
  title={Compose Your Policies! Improving Diffusion-based or Flow-based Robot Policies via Test-time Distribution-level Composition},
  author={Cao, Jiahang and Huang, Yize and Guo, Hanzhong and Zhang, Rui and Nan, Mu and Mai, Weijian and Wang, Jiaxu and Cheng, Hao and Sun, Jingkai and Han, Gang and others},
  journal={arXiv preprint arXiv:2510.01068},
  year={2025}
}

@article{wang2024poco,
  title={Poco: Policy composition from and for heterogeneous robot learning},
  author={Wang, Lirui and Zhao, Jialiang and Du, Yilun and Adelson, Edward H and Tedrake, Russ},
  journal={arXiv preprint arXiv:2402.02511},
  year={2024}
}

@inproceedings{xu2023xskill,
  title={Xskill: Cross embodiment skill discovery},
  author={Xu, Mengda and Xu, Zhenjia and Chi, Cheng and Veloso, Manuela and Song, Shuran},
  booktitle={Conference on robot learning},
  pages={3536--3555},
  year={2023},
  organization={PMLR}
}

@article{li2025towards,
  title={Towards Deploying VLA without Fine-Tuning: Plug-and-Play Inference-Time VLA Policy Steering via Embodied Evolutionary Diffusion},
  author={Li, Zhuo and Liu, Junjia and Dong, Zhipeng and Teng, Tao and Rouxel, Quentin and Caldwell, Darwin and Chen, Fei},
  journal={arXiv preprint arXiv:2511.14178},
  year={2025}
}

@article{wagenmaker2025steering,
  title={Steering your diffusion policy with latent space reinforcement learning},
  author={Wagenmaker, Andrew and Nakamoto, Mitsuhiko and Zhang, Yunchu and Park, Seohong and Yagoub, Waleed and Nagabandi, Anusha and Gupta, Abhishek and Levine, Sergey},
  journal={arXiv preprint arXiv:2506.15799},
  year={2025}
}

@inproceedings{wan2024contrastive,
  title={Contrastive region guidance: Improving grounding in vision-language models without training},
  author={Wan, David and Cho, Jaemin and Stengel-Eskin, Elias and Bansal, Mohit},
  booktitle={European Conference on Computer Vision},
  pages={198--215},
  year={2024},
  organization={Springer}
}

@inproceedings{jeong2025stylekeeper,
  title={StyleKeeper: Prevent Content Leakage using Negative Visual Query Guidance},
  author={Jeong, Jaeseok and Kim, Junho and Lee, Gayoung and Choi, Yunjey and Uh, Youngjung},
  booktitle={Proceedings of the IEEE/CVF International Conference on Computer Vision},
  pages={15760--15769},
  year={2025}
}

@article{li2024evaluating,
  title={Evaluating real-world robot manipulation policies in simulation},
  author={Li, Xuanlin and Hsu, Kyle and Gu, Jiayuan and Pertsch, Karl and Mees, Oier and Walke, Homer Rich and Fu, Chuyuan and Lunawat, Ishikaa and Sieh, Isabel and Kirmani, Sean and others},
  journal={arXiv preprint arXiv:2405.05941},
  year={2024}
}

@article{beyer2024paligemma,
  title={Paligemma: A versatile 3b vlm for transfer},
  author={Beyer, Lucas and Steiner, Andreas and Pinto, Andr{\'e} Susano and Kolesnikov, Alexander and Wang, Xiao and Salz, Daniel and Neumann, Maxim and Alabdulmohsin, Ibrahim and Tschannen, Michael and Bugliarello, Emanuele and others},
  journal={arXiv preprint arXiv:2407.07726},
  year={2024}
}
